\documentclass{article}
\usepackage[utf8]{inputenc} 
\usepackage[T1]{fontenc}    
\usepackage{hyperref}       
\usepackage{url}            
\usepackage{booktabs}       
\usepackage{nicefrac}       
\usepackage{microtype}      
\usepackage{amsfonts}       
\usepackage{amsmath}
\usepackage{amssymb}
\usepackage{paralist}
\usepackage{times}
\usepackage{subcaption}
\usepackage{graphicx}
\usepackage{graphics}
\usepackage{tabularx}
\usepackage{amsfonts,amsmath,amssymb,amsthm,url,xspace}
\usepackage{geometry}
\usepackage[numbers]{natbib}
\usepackage{lipsum}
\usepackage{setspace}
\usepackage{verbatim}
\usepackage{pifont}
\usepackage{authblk}
\usepackage{mathrsfs}
\usepackage{enumitem}
\usepackage[flushleft]{threeparttable}

\usepackage[colorinlistoftodos]{todonotes}
\usepackage{algorithm}
\usepackage{algpseudocode}

\usepackage[english]{babel}

\newtheorem{theorem}{Theorem}[section]

\newtheorem{lemma}[theorem]{Lemma}

\def\b0{{\boldsymbol{0}}}

\def\bx{{\boldsymbol{x}}}
\def\blambda{{\boldsymbol{\lambda}}}

\def\br{{\boldsymbol{r}}}

\def\bv{{\boldsymbol{v}}}
\def\bu{{\boldsymbol{u}}}

\def\bb{{\boldsymbol{b}}}

\def\by{{\boldsymbol{y}}}

\def\CDdual-ls{{\sf CDdual-ls}\xspace}

\def\CDdual{{\sf CDdual}\xspace}

\def\X{\mathcal{X}}
\def\S{\mathcal{S}}

\def\U{\mathcal{U}}

\def\CCDR1{{\sf CCD++}\xspace}

\def\e2006{{\sf TFIDF-2006}\xspace}

\def\rcv1{{\sf rcv1}\xspace}

\def\a9a{{\sf a9a}\xspace}

\def\ml1m{{\sf movielens1m}\xspace}

\def\movielens10m{{\sf movielens10m}\xspace}

\newcommand{\mR}{\mathbb{R}}
\newcommand{\mE}{\mathbb{E}}

\newcommand{\bw}{{\boldsymbol{w}}}

\newcommand{\Soft}{ {\rm Soft} }

\title{Scalable Peaceman-Rachford Splitting Method with Proximal Terms}
\author[1]{Sen Na\thanks{senna@uchicago.edu}}
\author[2]{Mingyuan Ma\thanks{mamingyuan@pku.edu.cn}}
\author[3]{Mladen Kolar\thanks{mkolar@chicagobooth.edu}}
\affil[1]{Department of Statistics, The University of Chicago}
\affil[2]{EECS, Peking University}
\affil[3]{Booth School of Business, The University of Chicago}

\date{\vspace{-5ex}}

\begin{document}	
	\maketitle

\begin{abstract}
Along with developing of Peaceman-Rachford Splittling  Method (PRSM), many batch algorithms based on it have been studied very deeply. But almost no algorithm focused on the performance of stochastic version of PRSM. In this paper, we propose a new stochastic algorithm based on PRSM, prove its convergence rate in ergodic sense, and test its performance on both artificial and real data. We show that our proposed algorithm, Stochastic Scalable PRSM (SS-PRSM), enjoys the $O(1/K)$ convergence rate, which is the same as those newest stochastic algorithms that  based on ADMM but faster than general Stochastic ADMM (which is $O(1/\sqrt{K})$). Our algorithm also owns wide flexibility, outperforms many state-of-the-art stochastic algorithms coming from ADMM, and has low memory cost in large-scale splitting optimization problems.
\end{abstract}

\section{Introduction}\label{Intro}

We consider the following minimization problem characterized by a
separable objective function with linear constraints:
\begin{align}\label{Mset}
\min_{\bx_1 \in \X_1, \bx_2 \in \X_2}& \theta_1(\bx_1)+\theta_2(\bx_2), \nonumber \\
\text{s.t.}\ \ \ &A\bx_1+B\bx_2=\bb,
\end{align}
where $A\in \mathbb{R}^{m\times n_1}$,
$B\in \mathbb{R}^{m\times n_2}$, $\bb\in \mathbb{R}^m$;
$\X_1\subset \mathbb{R}^{n_1}$ and $\X_2\subset \mR^{n_2}$ are
nonempty convex set; and $\theta_1: \X_1\rightarrow \mR$ and
$\theta_2: \X_2\rightarrow \mR$ are both convex functions.
The function $\theta_1(\cdot)$ is assumed to be of the form
$\theta_1(\cdot)=\frac{1}{n}\sum_{i=1}^{n}\theta_{1,i}(\cdot)$, where
$n$ is sample size and $\theta_{1,i}(\cdot)$ is the loss incurred on
$i$th sample. This setting is flexible enough to incorporate a number
of problems arising in machine learning and statistics, including Lasso, group Lasso
and logistic regression.


Many algorithms have been developed over the last thirty years to
solve the convex minimization problem in \eqref{Mset}, starting with the
Peaceman-Rachford splitting method (PRSM)~\cite{PRSM1955} and the Douglas-Rachford splitting method (DRSM)~\cite{DRSM1956}. Applying
DRSM to the dual of \eqref{Mset}, one gets the popular optimization
method called Alternating Direction Method of Multipliers
(ADMM)~\cite{glowinski1975ADMM, gabay1976dual, chan1978finite, boyd2011distributed},
which is fast in practice and easy to implement. Convergence rates, in
ergodic and non-ergodic sense, for ADMM have been studied recently.
For example,\cite{he20121,monteiro2013iteration,wang2013online} showed
ADMM has $O(1/K)$ ergodic rate, where $K$ stands for the number of
iterations, while \cite{davis2016convergence} established that
$O(1/\sqrt{K})$ non-ergodic convergence rate is tight for general
DRSM. Applying Nesterov's extrapolation to accelerate ADMM, one gets
$O(1/K)$ convergence rate~\cite{li2016optimal}. With additional
assumptions on the objective function, such as strongly convexity, we
the  convergence rate can be further strengthen~\cite{deng2016global,hong2017linear}.

In addition to all the theoretical developments, many new variants of
ADMM appeared, including both batch and stochastic versions of the
algorithm. \cite{ouyang2013stochastic} proposed a stochastic ADMM
iteration scheme and then showed its good performance on a large-scale
problem by using the first order approximation to the Lagrangian.
However, because of the noisy gradient and inexact approximation to
the stochastic function $\theta_1(\cdot)$, the stochastic ADMM can
only attain $O(1/\sqrt{K})$ ergodic convergence rate. Recently, a
number of accelerated stochastic version of ADMM that incorporate
variance reduction techniques (see ~\cite{johnson2013accelerating,
  defazio2014saga, zhao2015scalable, schmidt2017minimizing}) were
proposed with better convergence results --- SDCA-ADMM
\cite{suzuki2014stochastic}, SAG-ADMM
\cite{zhong2014fast}, SVRG-ADMM \cite{zheng2016fast}, and ASVRG-ADMM
\cite{liu2017accelerated}. SVRG-ADMM and SAG-ADMM enjoy $O(1/K)$
ergodic convergence rate and $O(1/\sqrt{K})$ non-ergodic convergence
rate. ASVRG-ADMM, which makes each iteration step expansive and needs
a large number of inner iterations, can have $O(1/K^2)$ ergodic rate
under general convex assumptions and a linear convergence rate under
strongly convex assumption. These developments effectively removed the
convergence rate gap between stochastic ADMM and batch ADMM.

On the other hand, development of PRSM and its variants is not as fast
as that of DRSM. Though PRSM always converges faster in experiments,
whenever it converges, the main difficulty for PRSM is that the
sequence generated by PRSM is not strictly contractive with respect to
the solution set of Problem~\eqref{Mset} \cite{he2014strictly}.
\cite{he2014strictly} proposed a method, called strictly contractive
PRSM (SC-PRSM), to overcome this difficulty by attaching an
underdetermined relaxation factor $\alpha\in(0,1)$ to the penalty
parameter $\beta$ when updating Lagrange multiplier.  After this
paper, mirroring the evolution of DRSM some new variants of SC-PRSM
have been developed. \cite{gu2015semi,na2016sparse} used two different
relaxation factors and showed the flexibility of this setting.
\cite{he2014strictly} showed SC-PRSM can attain the worst-case $O(1/K)$
convergence rate
in both ergodic and non-ergodic sense. With the exception of
\cite{na2016sparse}, development of stochastic algorithms based on
PRSM is lacking, even though it outperforms ADMM in many numerical
experiments. \cite{na2016sparse} developed an algorithm called
Stochastic Semi-Proximal-Based SC-PRSM (SSPB-SCPRSM), which contains
the Stochastic SC-PRSM as a special case, but with the convergence
rate of just $O(1/\sqrt{K})$ in ergodic sense, the same as stochastic
ADMM. Therefore the gap between batch PRSM and stochastic PRSM still
exists.

In this paper, we bridge the gap by developing a new accelerated
stochastic algorithm based on SC-PRSM, called Stochastic Scalable PRSM
(SS-PRSM). Compared to SC-PRSM, we use two different relaxation
factors, $\alpha$ and $\gamma$, to make it more flexible. We borrow
the general iteration structure from \cite{gu2015semi} and~\cite{na2016sparse}, but accelerate the iteration for $\bx_1$. This
adjustment will help us achieve $O(1/K)$ ergodic rate, improving the
$O(1/\sqrt{K})$ rate in~\cite{na2016sparse} and matching ADMM based
stochastic algorithms. Finally, we illustrate superiority over ADMM
based stochastic algorithms in numerical experiments, mirroring the
batch case. Our contribution in this paper are:
\begin{itemize}
\item Theoretically, we prove $O(1/K)$ ergodic convergence rate for
  the proposed algorithm. This bridges the ergodic convergence rate
  gap between stochastic PRSM and batch PRSM (note that the
  non-ergodic rate is still a open problem);
\item Comparing with related stochastic ADMM based algorithms, we add
  two proximal terms in iteration of $\bx_1$ and $\bx_2$, which improve
  flexibility;
\item Comparing with SVRG-ADMM~\cite{zheng2016fast}, we only
  accelerate $\bx_1$ iteration to get the same convergence rate;
\item Our algorithm is very flexible, leading to
  different new stochastic algorithms by setting $\alpha$, $\gamma$,
  $S$, $T$ properly;
\end{itemize}

The remainder of the paper is organized as follows. In
section~\ref{Back}, we will provide background, discuss related work
in more details, show fundamental iteration schemes, and provide
notations used throughout. In section~\ref{SVRG-SCPRSM}, we introduce
our algorithm. Theoretical convergence analysis is give in
section~\ref{Theorey}. Extensive numerical experiments to illustrate
the performance are in section~\ref{Numer}. Finally, section
~\ref{Con} concludes the paper and provides directions for future research.

{\bf Notations.} Throughout the papers, we will use the following notation:
\begin{itemize}
	\item $\forall \bx, \forall S\succeq 0$, define $\|\bx\|^2_S=\bx^TS\bx$; $D_{\X_1}=\sup_{\bx\in\X_1, \by\in\X_1}\|\bx-\by\|_2$;
	\item $\bu=\begin{pmatrix}
	\bx_1\\
	\bx_2
	\end{pmatrix}$; $\bw=\begin{pmatrix}
	\bx_1\\
	\bx_2\\
	\blambda
	\end{pmatrix}$; 
	$F(\bw)=\begin{pmatrix}
	-A^T\blambda\\
	-B^T\blambda\\
	A\bx_1+B\bx_2-\bb
	\end{pmatrix}$; 
	$\bar{\bw}_K=\begin{pmatrix}
	\bar{\bx}_{1K}\\
	\bar{\bx}_{2K}\\
	\bar{\blambda}_{K}\\
	\end{pmatrix}=\begin{pmatrix}
	\frac{1}{K}\sum_{k=1}^{K}\bx_1^{k+1}\\
	\frac{1}{K}\sum_{k=1}^{K}\bx_2^{k+1}\\
	\frac{1}{K}\sum_{k=1}^{K}\blambda^{k+1}\\
	\end{pmatrix}$;
	\item Define $\theta(\bu)=\theta_1(\bx_1)+\theta_2(\bx_2)$; given $v\in\mR^{p}$, $\Soft_{\kappa}(v)$ is vector valued soft-thresholding operator which is defined as $(\Soft_{\kappa}(v))_i=(1-\frac{\kappa}{|v_i|})_+v_i$, $\forall i=1,2,...,p$; define $\delta_i(\bx_1)=\theta'_{1i}(\bx_1)-\theta'_{1}(\bx_1)$ $\forall \bx_1\in\X_1, \forall i=1,...,n$.
\end{itemize}

\section{Background}\label{Back}

Modern data sets are getting ever larger, which drives the development
of efficient and scalable optimization algorithms. One promising
direction is the development of mini-batch and stochastic algorithms,
which can be thought of as a special case of mini-batching where one
sample point is involved in updating parameters. Another direction are
online algorithms that process data coming in streams
\cite{zinkevich2003online, hazan2008adaptive, wang2013online}. Our focus
in this paper, is developing of a scalable, stochastic algorithm for
solving large-scale optimization problems in \eqref{Mset}.  In a
stochastic algorithm, there is a trade off between computation speed
and the convergence speed. When a single sample is randomly chosen to
get an approximate descent direction, the computation is fast but the
convergence speed is slow. On the other hand, batch algorithms use all
the samples to find the exact descent direction, which results in
faster convergence rate, but the computation speed will be slower or
possibly unimplementable. Our proposed algorithm, balances the two
aspects by first computing the batch gradient, and then in an
iteration adjusts the gradient direction based on the current
sample. Our finial algorithm is comparable with a
batch algorithm like ADMM and SC-PRSM in ergodic sense. In addition, we add two pivotal proximal terms to make its implementation flexible.

\begin{algorithm}[t!]
	\caption{Stochastic ADMM}
	\label{SADMM}
    \begin{algorithmic}[1]
    	  \State Initialize $\bx_1^0$, $\bx_2^0$ and $\blambda^0$
    	  \For{$k=0,1,2,...$}
    	  \State $\bx_1^{k+1}=\arg\min_{\bx_1\in\X_1}L_{\beta,k}(\bx_1,\bx_2^k,\blambda^k)$;
    	  \State $\bx_2^{k+1}=\arg\min_{\bx_2\in\X_2}L_{\beta,k}(\bx_1^{k+1},\bx_2,\blambda^k)$;
    	  \State $\blambda^{k+1}=\blambda^k-\beta(A\bx_1^{k+1}+B\bx_2^{k+1}-\bb)$;
    	  \EndFor
  \end{algorithmic}
\end{algorithm}

We briefly introduce two fundamental stochastic iteration schemes that
are useful for development of our algorithm. First scheme is used in
the stochastic ADMM~\cite{ouyang2013stochastic}, where the noisy
gradient is used to approximate augmented Lagrangian as:
\begin{align}\label{aLag}
L_{\beta,k}(\bx_1,&\bx_2,\blambda)=\theta_1(\bx_1^k)+\langle \theta_1'(\bx_1^k,\xi_{k+1}),\bx_1\rangle \nonumber
+\theta_2(\bx_2) -\langle \blambda, A\bx_1+B\bx_2-\bb\rangle \nonumber
+\frac{\beta}{2}\|A\bx_1+B\bx_2-\bb\|^2+\frac{\|\bx_1-\bx_1^k\|^2}{2\eta_{k+1}}.
\end{align}
Here, $\xi_{k+1}$ can be seen as the selector of a sample that will be
used in computing a subgradient of $\theta_1(\cdot)$ in the $k$th
iteration.  In our setting
$\theta_1(\cdot)=\frac{1}{n}\sum_{i=1}^{n}\theta_{1,i}(\cdot)$, and we
can take $\xi_{k}\stackrel{iid}{\sim}\U\{1,n\}$\footnote{We use
	$\U\{1,n\}$ to denote a uniform distribution among discrete indices
	$1,2,...,n$.}.
More generally, if
$\theta_1(\cdot)=\mathbb{E}_{\xi}\theta_1(\cdot,\xi)$, we can take
$\xi_{k}\stackrel{iid}{\sim}P$ for some fixed distribution
$P$. Furthermore, $\blambda$ is the Lagrangian multiplier, $\beta$ is
the predefined penalty parameter, and $\eta_{k+1}$ is the time-varying
step size. As iteration number goes up, we expect to find a triplet
$(\bx_1^*,\bx_2^*,\blambda^*)$ such that $\forall \bx_1\in \X_1$,
$\forall \bx_2\in \X_2$, $\forall \blambda\in\mR^m$, we have
\begin{equation*}
L_{\beta,k}(\bx_1^*,\bx_2^*,\blambda)\leq L_{\beta,k}(\bx_1^*,\bx_2^*,\blambda^*)\leq L_{\beta,k}(\bx_1,\bx_2,\blambda^*).
\end{equation*}
Details of stochastic ADMM are given in Algorithm ~\ref{SADMM}, which
successively updates $\bx_1$, $\bx_2$, $\blambda$.

\begin{algorithm}[t!]
	\caption{Stochastic SPB-SCPRSM}
	\label{S-SPB-SCPRSM}
    \begin{algorithmic}[1]
    	  \State Initialize $\bx_1^0$, $\bx_2^0$ and $\blambda^0$
    	  \For{$k=0,1,2,...$}
    	  \State $\bx_1^{k+1}=\arg\min_{\bx_1\in\X_1}L_{\beta,k}(\bx_1,\bx_2^k,\blambda^k)+\frac{1}{2}\|\bx_1-\bx_1^k\|_S^2$;
    	  \State $\blambda^{k+1/2}=\blambda^k-\alpha\beta(A\bx_1^{k+1}+B\bx_2^{k}-\bb)$;
    	  \State $\bx_2^{k+1}=\arg\min_{\bx_2\in\X_2}L_{\beta,k}(\bx_1^{k+1},\bx_2,\blambda^{k+1/2})+\frac{1}{2}\|\bx_2-\bx_2^{k}\|_T^2$;
    	  \State $\blambda^{k+1}=\blambda^{k+1/2}-\gamma\beta(A\bx_1^{k+1}+B\bx_2^{k+1}-\bb)$;
    	  \EndFor
  \end{algorithmic}
\end{algorithm}

The second iteration scheme is used in SSPB-SCPRSM \cite{na2016sparse}
and is shown in Algorithm~\ref{S-SPB-SCPRSM}.  The additional
parameters $\alpha,\gamma,S,T$ make Algorithm~\ref{S-SPB-SCPRSM} more
flexible.  The feasible range for $\alpha$ and $\gamma$ is
$[0,1)\times(0,1]$ and $S,T\succeq0$. Setting $S=T=0$ results in
stochastic SCPRSM. When $\alpha=\gamma$ goes to $1$, the algorithm
result in stochastic PRSM. Different from stochastic ADMM, it creates
an intermediate iterate $\blambda^{k+1/2}$ between $\bx_1$ and
$\bx_2$, which always appears in PRSM based algorithms.
Without additional parameters $\alpha, \gamma$, this step can still make PRSM outperform ADMM in experiment but also results in the
loss of property of strict contraction~\cite{he2014strictly,gu2015semi}.

\section{Stochastic Scalable PRSM}\label{SVRG-SCPRSM}

In this section, we describe our proposed stochastic scalable PRSM
algorithm, which improves over SC-PRSM. A gradient estimate based on a
single sample has a large variance and one requires the time-varying
step size to decay to zero in order to ensure convergence. The order
of this decay is essential for obtaining good convergence rate
\cite{na2016sparse}. Here, we use a variance reduction trick (inspired
by~\cite{johnson2013accelerating}) to incorporate information from all
the samples in a stochastic setting.

Instead of using a first order approximated Lagrangian, we minimize
the following augmented Lagrangian:
\begin{align}\label{augL}
L_{\beta}(\bx_1,\bx_2,\blambda)=&\theta_1(\bx_1)+\theta_2(\bx_2)-\langle \blambda, A\bx_1+B\bx_2-\bb\rangle \nonumber
+\frac{\beta}{2}\|A\bx_1+B\bx_2-\bb\|^2.
\end{align}
Compared to Algorithm~\ref{S-SPB-SCPRSM}, SS-PRSM only adjusts the iteration for
$\bx_1$. Suppose at iteration $k$, we have.
$\bx_1^k, \bx_2^k, \blambda^k$. We define a gradient function
$G_{\beta}^k(\bx_1)$ associated with $\bx_1$ as
\begin{equation}
G_{\beta}^k(\bx_1)=L_{\beta}(\bx_1,\bx_2^k,\blambda^k)+\frac{1}{2}\|\bx_1-\bx_1^k\|_S^2,
\end{equation}
where $S\succeq0$ is a predefined matrix. Note that if $S\succ0$, the
proximal term $\|\bx_1-\bx_1^k\|^2_S$ has the same function as
$\frac{\|\bx_1-\bx_1^k\|^2}{2\eta_{k+1}}$ term in \eqref{aLag}, but we
do not need $\eta_k\rightarrow0$. Note that we can use $S_k$, which
changes with iterations, but for simplicity and implementability we
will fix $S$. In our algorithm, stochastic variance reduced gradient descent is used to
minimize $G_{\beta}^k(\bx_1)$. Algorithm~\ref{SVRG} summarizes the steps used to update $\bx_1$.

\begin{algorithm}[t]
	\caption{SS-PRSM (one step)}
	\label{SVRG}
	\begin{algorithmic}[1]
\State Specify $M_k$, $\eta_k$;
\State Set $\tilde{\bx}_1=\bx_1^k$; $\bx_{1,0}=\tilde{\bx}_1$;
\State Compute batch gradient $\tilde{\mu}=(G_{\beta}^{k})^{\prime}(\tilde{\bx}_1)$;
\For {$t=1,2,...M_k-1$}
\State Randomly draw $\xi^t_{k+1}$ from $\U\{1,n\}$;
\State $\nabla g_{\beta}(\bx_{1,t-1},\xi^t_{k+1})=(G_{\beta}^{k})^{\prime}(\bx_{1,t-1},\xi^t_{k+1})-(G_{\beta}^{k})^{\prime}(\tilde{\bx}_1,\xi^t_{k+1})+\tilde{\mu}$;
\State $\bx_{1,t}=\bx_{1,t-1}-\eta_k\nabla g_{\beta}(\bx_{1,t-1},\xi^t_{k+1})$;
\EndFor
\State $\bx_1^{k+1}=\frac{1}{M_k}\sum_{t=0}^{M_k-1}\bx_{1,t}$.
	\end{algorithmic}
\end{algorithm}

The iteration scheme for $\bx_2$ and $\blambda$ are the same as
Algorithm~\ref{S-SPB-SCPRSM} (lines 4-6). Note that using
$L_{\beta,k}(\cdot)$ or $L_\beta(\cdot)$ for updating $\bx_2$ will
result in the same update. The term
$\nabla g_{\beta}(\bx_{1,t-1},\xi_{k+1}^t)$ is crucial here because it
involves $\tilde{\mu}$, which collects the "information" we get from
other samples that are not selected in the stochastic update.  With
this choice of $\nabla g_{\beta}(\bx_{1,t-1},\xi_{k+1}^t)$ we have
that
$\mathbb{E}(\bx_{1,t}|\bx_{1,t-1})=\bx_{1,t-1}-\eta_k(G_{\beta}^k)'(\bx_{1,t-1})$.
Finally, we note that saving all $\bx_{1,t}$ from 0 to $M_k-1$ is not
necessary, as an incremental vector can be used to obtain $\bx_{1}^{k+1}$.

Computationally, one iteration of SS-PRSM is slower than one iteration
in Algorithm~\ref{SADMM} and~\ref{S-SPB-SCPRSM}, because we still need
to compute the batch gradient. However, the inner iteration is fast
and comparable with that of ADMM based algorithms~\cite{zhong2014fast,
	zheng2016fast, liu2017accelerated}, while the overall number of
iterations is lower. Also, from our algorithm, we see only accelerate one variable is enough to attain $O(1/t)$ rate instead of involving $\bx_2$ and Lagrangian multiplier $\blambda$. In the following section, we will show our main theoretical result.

\section{Convergence Analysis}\label{Theorey}

In this section, we study the convergence rate for SS-PRSM and
establish $O(1/K)$ ergodic rate. To measure the convergence rate, we
will use following criterion function which is from variational inequality:
$\mE[\theta(\bar{\bu}_K)-\theta(\bu^*)+\langle\bar{\bw}_K-\bw, F(\bw)\rangle]$. We will show that $\forall \bw\in\Omega$ where $\Omega=\S\times\mR^m$ and $\S=\{(\bx_1,\bx_2): A\bx_1+B\bx_2=b, \bx_1\in \X_1, \bx_2\in \X_2\}$, we have
\begin{equation}
\mE \bigg[\theta(\bar{\bu}_K)-\theta(\bu^*)+\langle\bar{\bw}_K-\bw, F(\bw)\rangle\bigg]\approx O(\frac{1}{K}).
\end{equation}
The above criterion is equivalent to
$\mE[\theta(\bar{\bu}_K)-\theta(\bu^*)+\rho\|A\bar{\bx}_{1K}+B\bar{\bx}_{2K}-\bb\|_2]$
for $\forall \rho>0$, which was used in
~\cite{ouyang2013stochastic, zhao2015scalable, liu2017accelerated}.
Furthermore, since our algorithm is only modifying the updating scheme
for $\bx_1$, we will borrow part of arguments given
in~\cite{na2016sparse}. We start by listing our fairly mild
assumptions:

\noindent\textbf{A1} $\theta_{1}(\cdot)$ and $\theta_2(\cdot)$ are two convex function with $\theta_{1}(\cdot)=\frac{1}{n}\sum_{i=1}^{n}\theta_{1i}$.

\noindent\textbf{A2} For all $i$, $\theta_{1i}$ is $\nu_i$-smooth ---
$\forall \bx, \by\in \X_1$,
$\|\theta_{1i}'(\bx)-\theta_{1i}'(\by)\|\leq
\nu_i\|\bx-\by\|.$
 Furthermore, we let $\nu=\max_{i}\nu_i$.

\noindent\textbf{A3} We assume $D_{\X_1}<\infty$.  Otherwise, we only
solve a minimization in a large enough but finite space.

Note that the assumptions are quite general. In particular, we do not
need the smoothness on $\theta_2(\cdot)$. \textbf{A3} is
necessary for all ADMM and PRSM method to guarantee convergence.

Next, we present two lemmas which will be used to prove the main
result. The first lemma measure the decrease for $\bx_1$ in one step.

\begin{lemma}\label{Lem4.1}
	For $k=0,1,2...$ fixed, suppose we have $\bx_1^k, \bx_2^k, \blambda^k$ (so $\tilde{\bx}_1=\bx_1^k$), then after getting $\bx_{1}^{k+1}$ by doing SS-PRSM for one step we will have $\forall \bx_1\in \X_1$,
	\begin{align*}
	&\theta_1(\bx_1)-\theta_1(\bx_1^{k+1})-\frac{1}{M_k}\sum_{t=0}^{M_k-1}\Delta(k,t)+\frac{D_{\X_1}^2}{2\eta_kM_k}
	+\frac{\eta_k}{2M_k}\sum_{t=0}^{M_k-1}\|\nabla g_{\beta}(\bx_{1,t},\xi_{k+1}^{t+1})\|^2\\ 
&+\langle \bx_1-\bx_1^{k+1},S(\bx_{1}^{k+1}-\bx_1^k)-A^T\blambda^k+\beta A^T(A\bx_{1}^{k+1}+B\bx_2^k-\bb)\rangle\geq 0,
	\end{align*}
	where $\Delta(k,t)=\langle \delta_{\xi_{k+1}^{t+1}}(\bx_{1,t})-\delta_{\xi_{k+1}^{t+1}}(\tilde{\bx}_{1}), \bx_{1,t}-\bx_1\rangle$.
\end{lemma}
Let focus on the term $\frac{1}{M_k}\sum_{t=0}^{M_k-1}\|\nabla g_{\beta}(\bx_{1,t},\xi_{k+1}^{t+1})\|^2$ on the left hand side, it's an approximation to $\mE[\|\nabla g_{\beta}(\bx_{1,t},\xi_{k+1}^{t+1})\|^2]$. To show the lower bound for decrease of $\bx_1$, we need an upper bound for this expectation. Following lemma gives this upper bound.

\begin{lemma}\label{Lem4.2}
	Define $C_k=\|(G_{\beta}^{k})^{\prime}(\bx_1^k)\|$ for $k=0,1,2...$, then $\forall t=0,2,...,M_k-1$ we have
	\begin{align*}
	\mE [\|\nabla g_{\beta}(\bx_{1,t},\xi_{k+1}^{t+1})\|^2]\leq 2D^2_{\X_1}(\nu+\beta\sigma^2_A+\sigma_S)^2+2C_k^2.
	\end{align*}
\end{lemma}
Based on these two lemmas, we are ready to present our main result. However, our inner iteration step-size $\eta_k$ and iteration number is pivotal. We will show in the prove that we can simply set $M_k = C + C_k^2$ and $\eta_k = \frac{1}{M_k k^2}$ for some constant $C$ to attain global convergence. Because $C_k$ will converges to 0 as $k$ increasing, we see fewer and fewer inner iterations we will need as algorithm progresses.

\begin{theorem}\label{thm1}
	Let the sequence $\{\bw^{k}\}_{k=1}^{\infty}$ be generated by SS-PRSM, under above setting for $\eta_k$ and $M_k$ and let $\alpha_k\in[0,1)$ and $\gamma\in(0,\frac{1-\alpha+\sqrt{(1+\alpha)^2+4(1-\alpha^2)}}{2})$, we will have $\forall \bw\in\Omega$
	\begin{align*}
	\mE[\theta(\bar{\bu}_K)-\theta(\bu)+\langle \bar{\bw}_K-\bw, F(\bw)\rangle]\approx O(\frac{1}{K}).
	\end{align*}
	Further, $\forall \rho>0$, we have
	\begin{align*}
	\mE[\theta(\bar{\bu}_K)-\theta(\bu^*)+\rho\|A\bar{\bx}_{1K}+B\bar{\bx}_{2K}-\bb\|_2]\approx O(\frac{1}{K}).
	\end{align*}
	where $\bu^*=\begin{pmatrix}
	\bx_1^*\\
	\bx_2^*
	\end{pmatrix}$ is true solution.
\end{theorem}

We give a brief proof sketch. First, because our iteration scheme for $\bx_2$ and $\blambda$ are the same as~\cite{na2016sparse}, we modified their results to fit it our algorithm. Combining their part of results with our Lemma~\ref{Lem4.1}, we have following argument to measure the total decrease of criteria function. For self-contain, we have showed how to prove this argument in our supplementary material.

{\bf Argument:} Let the sequence $\{\bw^k\}_{k=1}^{\infty}$ be generated by SS-PRSM. If $\alpha\in[0,1)$ and $\gamma\in(0,\frac{1-\alpha+\sqrt{(1+\alpha)^2+4(1-\alpha^2)}}{2})$, then $\forall \bw\in \Omega$ we have
\begin{align*}
&\theta(\bu^{k+1})-\theta(\bu)+\langle \bw^{k+1}-\bw, F(\bw^{k+1})\rangle\leq\frac{D_{\X_1}^2}{2\eta_kM_k}
-\frac{1}{M_k}\sum_{t=0}^{M_k-1}\Delta(k,t)+\frac{\eta_k}{2M_k}\sum_{t=0}^{M_k-1}\|\nabla g_{\beta}(\bx_{1,t},\xi_{k+1}^{t+1})\|^2\\
&+\frac{1}{2}(\|\bw^k-\bw\|^2_G+\zeta_{\alpha}\|\bx_2^k-\bx_2^{k-1}\|_T^2+\rho_{\alpha,\gamma}\beta\|\br^k\|^2)
-\frac{1}{2}(\|\bw^{k+1}-\bw\|^2_G+\zeta_{\alpha}\|\bx_2^{k+1}-\bx_2^{k}\|_T^2+\rho_{\alpha,\gamma}\beta\|\br^{k+1}\|^2),
\end{align*}
where $\zeta_\alpha, \rho_{\alpha,\gamma}\geq 0$, $\br^{k}=A\bx_1^{k}+B\bx_2^k-\bb$ for $k=1,2,...$, and $G
=\left( \begin{array}{ccc}
S & 0 &0 \\
0 & T+\frac{\alpha+\gamma-\alpha\gamma}{\alpha+\gamma}\beta B^{T}B & -\frac{\alpha}{\alpha+\gamma}B^T \\
0 & -\frac{\alpha}{\alpha+\gamma}B & \frac{1}{(\alpha+\gamma)\beta}I_m \end{array} \right)$.

Note that $\mE[\Delta(k,t)]=0$. So summing from $1$ to $K$ and taking expectation on both side, we can show
\begin{align*}
\mE[\theta(\bar{\bu}_K)-\theta(\bu)+\langle \bar{\bw}_K-\bw, F(\bw)\rangle]
&\leq \frac{1}{2K}(\|\bw^1-\bw\|^2_G+\zeta_{\alpha}\|\bx_2^1-\bx_2^{0}\|_T^2+\rho_{\alpha,\gamma}\beta\|\br^1\|^2)\\
&+\frac{1}{K}\sum_{k=1}^{K}\bigg(\frac{D_{\X_1}^2}{2\eta_kM_k}+\eta_k(\underbrace{D^2_{\X_1}(\nu+\beta\sigma^2_A+\sigma_S)^2}_{C^2}+C_k^2)\bigg).
\end{align*}
So, under our setting for $\eta_k$ and $M_k$, we will get our main theorem. For the second part of the theorem, because $\bw$ is random in above inequality, we consider $\bw$ to be $\tilde{\bw}=(\bx_1^*, \bx_2^*, \blambda)$ where $\blambda\in\mathcal{B}_{\rho}=\{\blambda, \|\blambda\|_2\leq\rho\}$ for any $\rho>0$. Plug into above inequality and using following fact
\begin{align*}
\theta(\bar{\bu}_K)-\theta(\bu^*)+\langle \bar{\bw}_K-\tilde{\bw}, F(\tilde{\bw})\rangle
=\theta(\bar{\bu}_K)-\theta(\bu^*)-\blambda(A\bar{\bx}_{1K}+B\bar{\bx}_{2K}-\bb),
\end{align*}
we can show SS-PRSM convergence order in the second criteria. The detailed proof is proposed in supplementary material.

From above theorem, we see SS-PRSM has $O(1/K)$ ergodic convergence rate. So, we have showed that only adjusting the iteration in $\bx_1$, we can gain the same convergence rate as most of ADMM based stochastic algorithm like SVRG-ADMM (see ~\cite{zheng2016fast}).

\section{Numerical Experiments}\label{Numer}

In this section, we run extensive numerical experiments to test
performance of our algorithm. Since the goal is to bridge the
convergence rate gap between stochastic and batch algorithms based on
PRSM, we will also use SC-PRSM as one of competitors. Specifically, we
use the following algorithms for comparison:
\begin{enumerate}[topsep=1pt,itemsep=-1ex,partopsep=1ex,parsep=1ex]
	\item stochastic algorithms:
	\begin{itemize}[topsep=0.5pt,itemsep=-1ex,partopsep=1ex,parsep=1ex]
        \item PRSM-based: SS-PRSM (our proposed algorithm);
          SSPB-SCPRSM \cite{na2016sparse};
        \item ADMM-based: SCAS-ADMM \cite{zhao2015scalable};
          SVRG-ADMM \cite{zheng2016fast}; S-ADMM \cite{ouyang2013stochastic};
	\end{itemize}
      \item batch algorithms: SC-PRSM \cite{he2014strictly}; ADMM
        \cite{glowinski1975ADMM,chan1978finite,gabay1976dual,boyd2011distributed}.
\end{enumerate}
We use both simulated data and real data and focus on three
optimization problems arising in machine learning: Lasso, group Lasso,
and sparse logistic regression. These problems can be easily
formulated as an optimization programs of the form \eqref{Mset}, as we
show below. We try a wide range of settings for our parameters. We use
$X=\begin{pmatrix} X_1 & X_2 & \cdots & X_n
\end{pmatrix}^T\in\mR^{n\times p}$
to denote the sample matrix where $p$ is the number of predictors;
$Y=\begin{pmatrix} Y_1 & Y_2 & \cdots & Y_n
\end{pmatrix}\in\mR^n$
is the response vector; $y$ and $x$ denote one row of $Y$ and $X$
respectively; $Z\in\mR^p$ is the target parameter; $\S={\rm supp}(Z)$
and $|\S|$ is the number of nonzero coefficients. Simulation results
are averaged over $20$ independent runs.


\subsection{Lasso}

\begin{figure}[h]
	\centering
	\begin{subfigure}[b]{0.4\textwidth}
		\includegraphics[width=\textwidth]{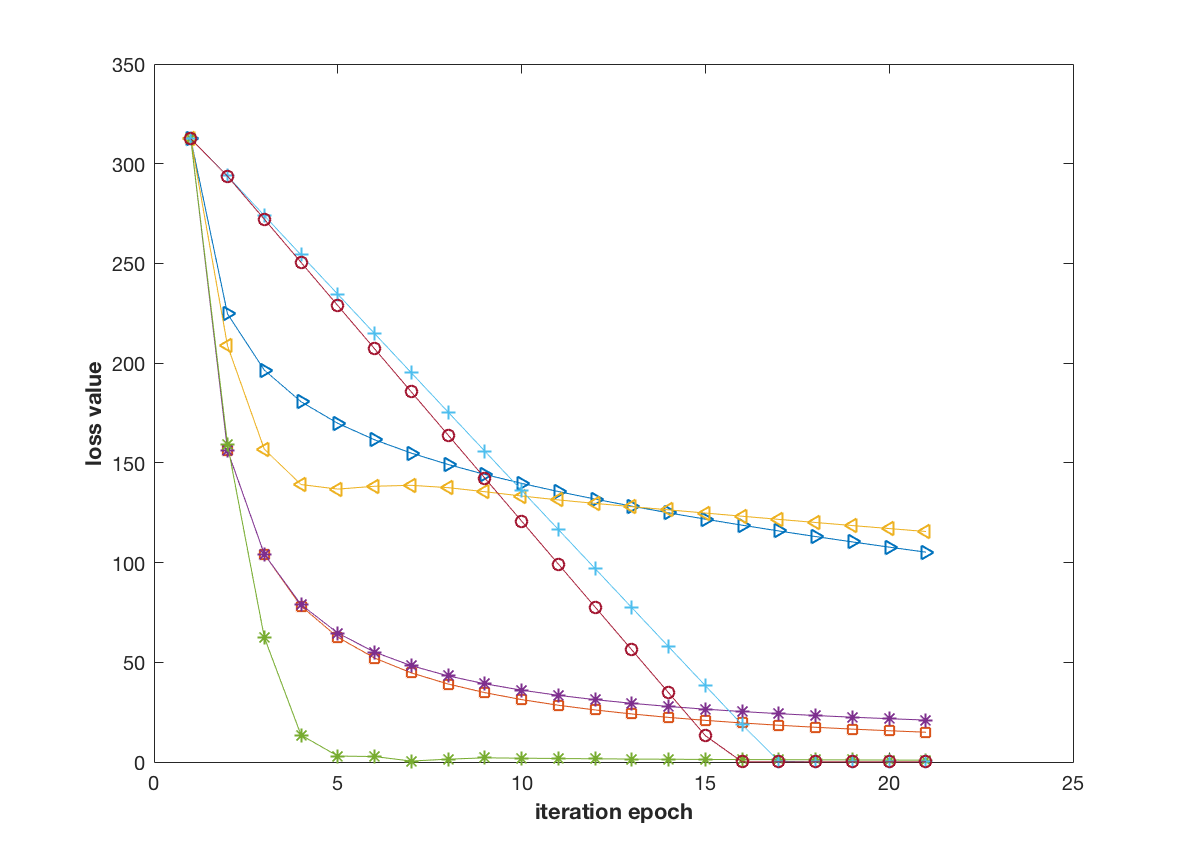}
		\caption{Lasso}
		\centering
		\label{Lasso1}
	\end{subfigure}
	~
	\begin{subfigure}[b]{0.4\textwidth}
		\includegraphics[width=\textwidth]{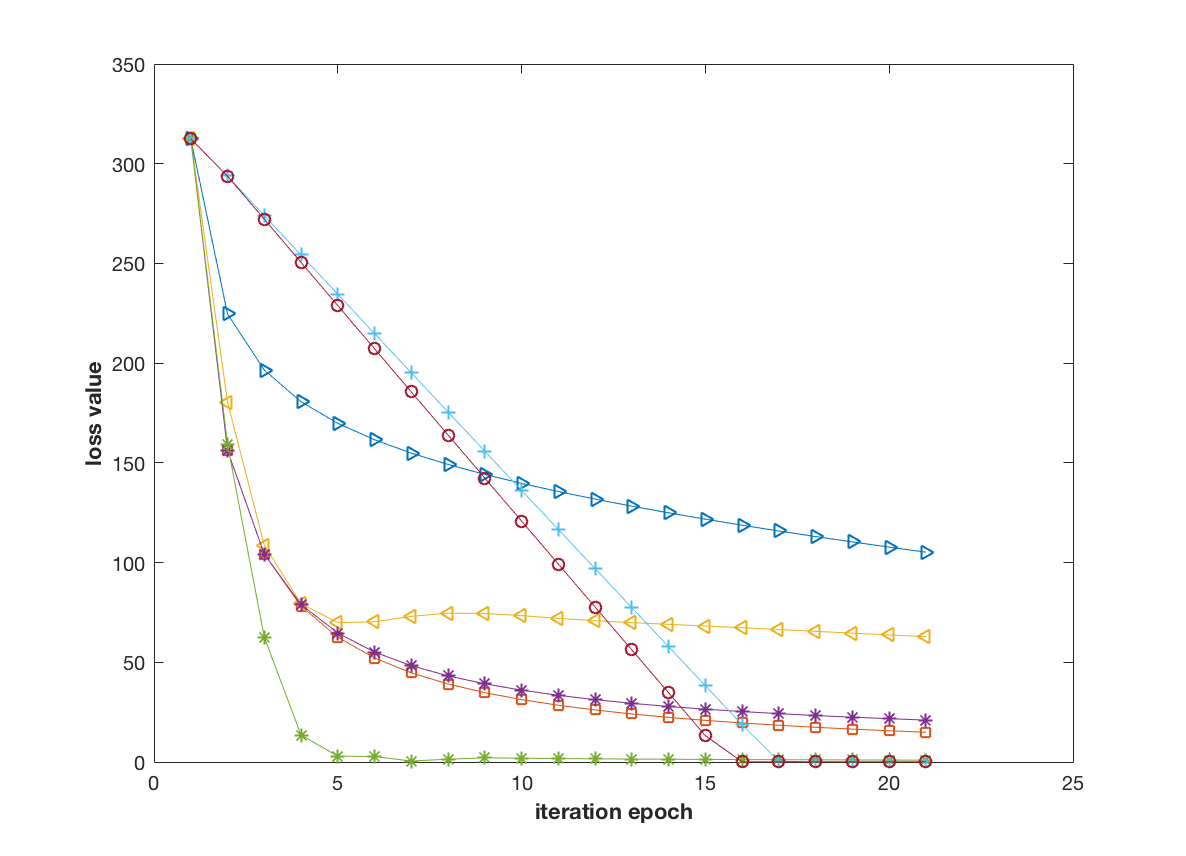}
		\caption{Lasso}\centering
		\label{Lasso2}
		\centering
	\end{subfigure}
	\\
	\begin{subfigure}[b]{0.4\textwidth}
		\includegraphics[width=\textwidth]{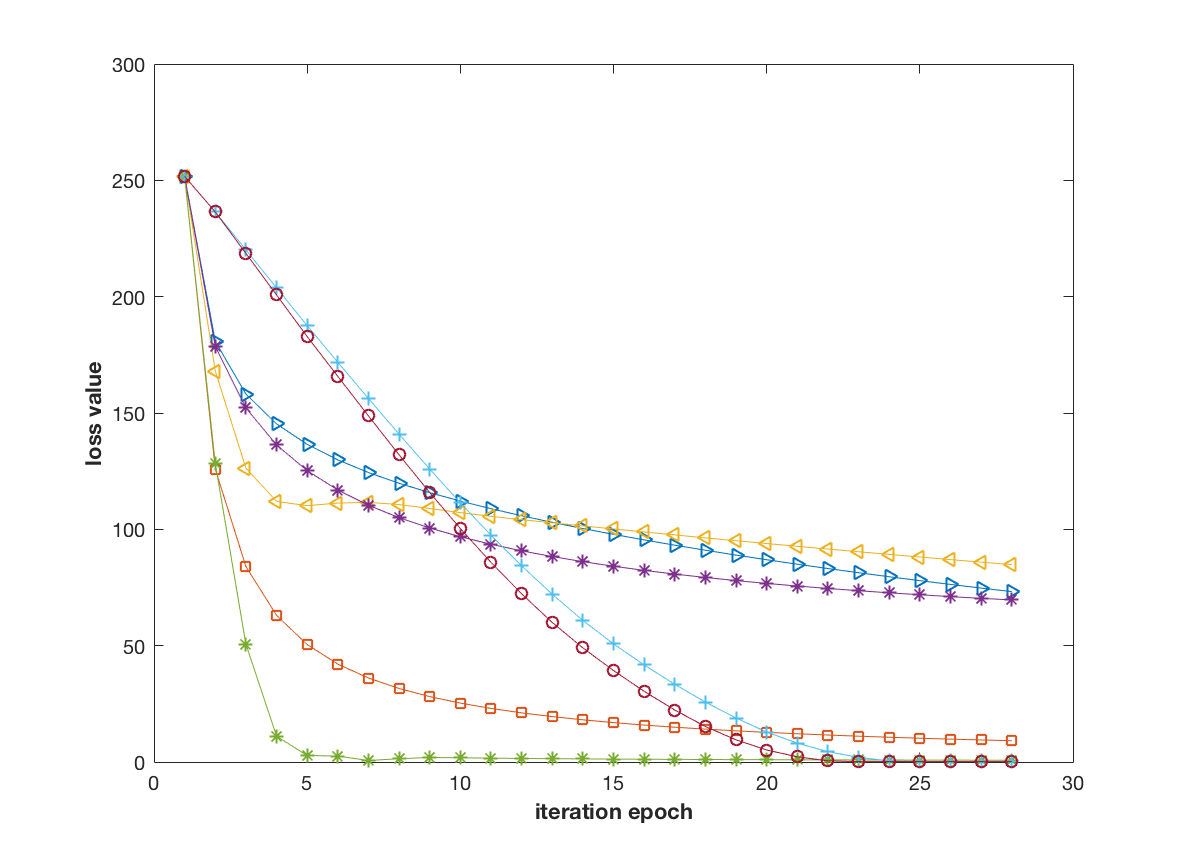}
		\caption{group Lasso}\centering
		\label{groupLasso}
		\centering
	\end{subfigure}
~
	\begin{subfigure}[b]{0.4\textwidth}
	\includegraphics[width=\textwidth]{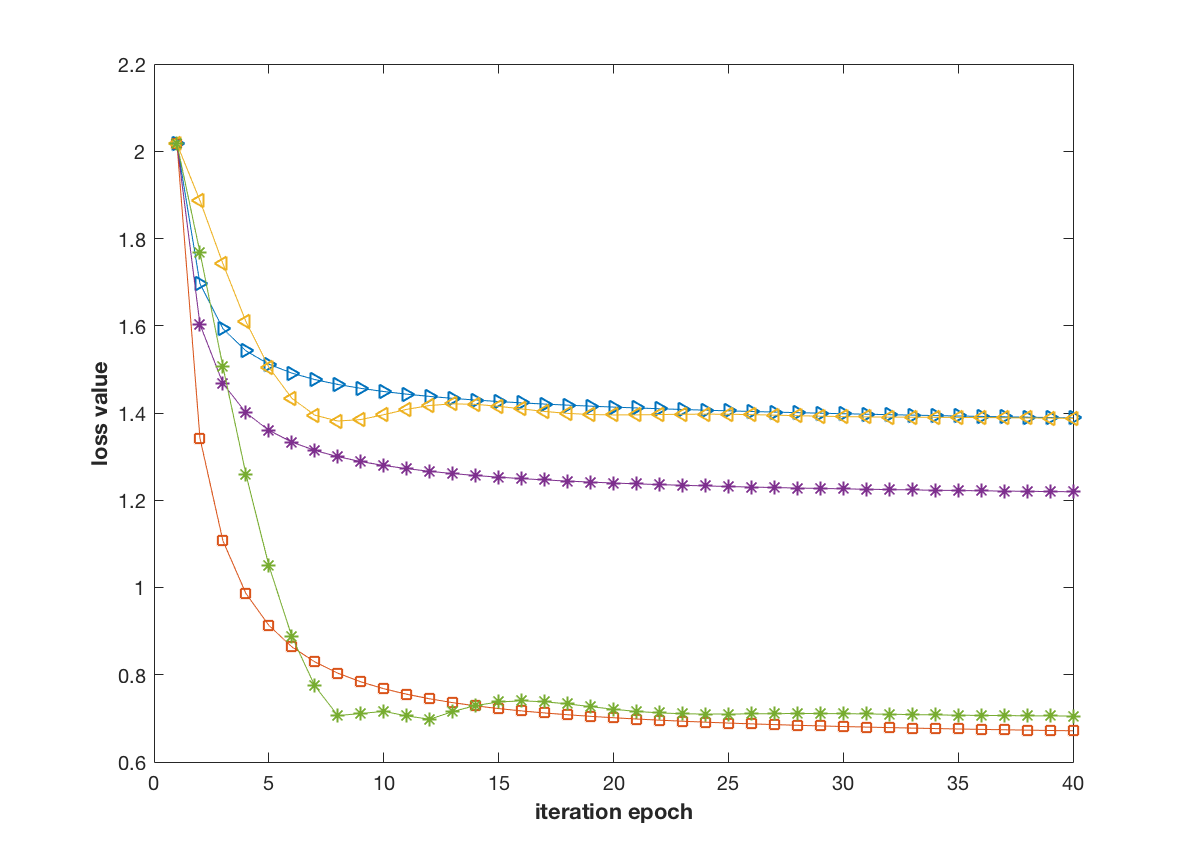}
	\caption{sparse logistic}\centering
	\label{LogisS}
	\centering
\end{subfigure}
	\includegraphics[width=15cm,height=0.27cm]{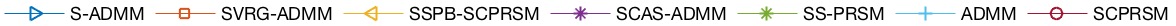}
	\caption{\textit{This figure shows loss decay trajectory for
            different methods on simulated data sets. The left plot is
            the lasso problem with $S=I$ and the second plot is lasso
            with $S = 5I$. We see that both SS-PRSM and SSPB-PRSM can
            be influenced by how one sets $S$. However, because of
            inner iteration for $\bx_1$ in SS-PRSM, it is robust to
            the choice of $S$. We do not implement SC-PRSM and ADMM
            for sparse logistic regression case because there is no
            closed form for subproblem (one usually use the
            Newton's method to solve it).}}\label{fig:1}
\end{figure}

Lasso problem can be formulated as
\begin{align*}
\min_{Z_1,Z_2} &\frac{1}{n}\|Y-XZ_1\|_2^2+\zeta\|Z_2\|_1\\
\text{s.t.}\ \ \ &Z_1-Z_2=0.
\end{align*}
We set $n=2000$, $p=5000$, and $|\S|=200$. We construct $X$ by drawing
each entry independently from $N(0,1)$ and then normalizing each row
as $x_i=\frac{x_i}{\|x_i\|_2}, \forall i=1,2,...,n)$. The parameter
$Z$ is generated by uniformly drawing $|\S|$ indices from 1 to $p$
setting $Z_{\S}\sim N(0, I_{|\S|})$. Finally, $Y=XZ+\epsilon$ where
$\epsilon\sim N(0,0.01I_n)$. The regularization parameter $\zeta$ is
set as $\zeta=0.1\|XZ\|_{\infty}$. Since $Z$ is sparse we assume
$D_{\X}\approx |\S|$ when setting $\eta_k$ and $M_k$. The setting for
other parameters is tabulated in Table~\ref{Parset}. We only consider
setting $T=a\cdot I$, while $S$ can be set arbitrarily. This is
because other forms of $T$ might make the subproblem for finding $Z_2$
hard to solve. The detailed iteration scheme is given in
Algorithm~\ref{SVRGLasso} in supplementary material. The loss value
decay plot is shown in Figure~\ref{Lasso1} and Figure~\ref{Lasso2}.
From the plot, we see that S-ADMM and SSPB-SCPRSM require many more
passes over the data. Our SS-PRSM is competitive with the
state-of-the-art stochastic ADMM based algorithm.
\begin{table}
	\centering
	\begin{tabular}{ |c|c|c|c|c|c|c| }
		\hline
		Model & $\epsilon$ & $\beta$ &$\alpha$& $\gamma$ & $S$ & $T$\\
		\hline
		Lasso & $10^{-11}$ & 1 & 0.9 & 0.1 & $I$ & $I$\\
		\hline
Lasso & $10^{-11}$ & 1 & 0.9 & 0.1 & $5I$ & $I$\\
\hline
group Lasso & $10^{-11}$ & 1 & 0.9 & 0.1 & $I$ & $I$\\
\hline
sparse logistic & $10^{-8}$ & 1 & 0.5 & 0.3 & $3I$ & $3I$\\
\hline
	\end{tabular}
	\caption{parameter setting}\label{Parset}
\end{table}

\subsection{Group Lasso}

Group Lasso problem can be formulated as
\begin{align*}
\min_{Z_1,Z_2} &\frac{1}{n}\|Y-XZ_1\|_2^2+\zeta\sum_{i=1}^{N}\|Z_{2i}\|_2\\
\text{s.t.}\ \ \ &Z_{1i}-Z_{2i}=0, i=1,2,...,N,
\end{align*}
where $N$ is the number of groups; $Z_{1i}$ and $Z_{2i}\in\mR^{n_i}$
is the parameter vector in each group. Let
$Z_1=\begin{pmatrix} Z_{11} & \cdots & Z_{1N}\end{pmatrix}^T$.

We set $n=3000$, $N=300$, $n_i\stackrel{iid}{\sim}\U\{1,30\}$, and
$p=\sum_{i=1}^{N}n_i$. $X$ is created as in the Lasso case. $Z[i]$
denotes the parameter vector in the $i$th group. We let
$|S_i|=\lfloor 0.15n_i\rfloor$ and $Z[i]_j\stackrel{iid}{\sim}N(0,1)$
for $j\in S_i$. Finally, $Y=XZ+\epsilon$ with
$\epsilon\sim N(0,0.01I_n)$ and
$\zeta=0.1\max\{\|XZ[1]\|_{2},...,\|XZ[N]\|_{2}\}$. The iteration
scheme is the same as in the Lasso case except that the $16$th row of
Algorithm~\ref{SVRGLasso} (see supplement) is replaced by
\begin{align*}
Z_{2i}^{t+1}=\Soft_{\frac{\zeta}{a+\beta}}\left(\frac{aZ_{2i}^t+\beta Z_{1i}^{t+1}-\blambda^{t+1/2}_i}{a+\beta}\right), i=1,2,...,N,
\end{align*}
where $Z_{1i}^{t+1}$ (different from $Z_{1,i}$ in updating $Z_1$) and
$\blambda_i^{t+1}$ are parameter and Lagrange multiplier for the $i$th
group, respectively. Here,
$\Soft_{\kappa}(v)=(1-{\kappa}/{\|v\|_2})_+v$ with
$\Soft_{\kappa}(0)=0$. The loss decay plot is in Figure
~\ref{groupLasso}. We see that SS-PRSM and SVRG-ADMM are the two
fastest algorithms, with SS-PRSM being slightly better than
SVRG-ADMM. We also note the slow convergence rate of SSPB-SCPRSM and S-ADMM.

\subsection{Sparse logistic regression}

Sparse logistic regression can be formulated as
\begin{align*}
\min_{Z_1,Z_2} &\frac{1}{n}\sum_{i=1}^{n}\log(1+\exp(-Y_iX_i^TZ_1))+\zeta\|Z_2\|_1\\
\text{s.t.}\ \ \ &Z_1-Z_2=0.
\end{align*}
We set $n=100$, $p=400$, and $|\S|=100$. Sample matrix $X$ is
generated as follows: each sample point $x_i$ has twenty nonzero
entries independently drawn from $N(0,1)$ with the index generated
uniformly from 1 to $p$; $Z$ is simulated as in the Lasso case and
$Y=\text{sign}(XZ+\epsilon)$ where $\epsilon\sim N(0,0.01I_n)$ and
$\text{sign}(\cdot)$ is computed entry-wise. For the regularization
parameter we use the setting in~\cite{koh2007interior} where
$\zeta=\frac{0.1}{n}\|\sum_{Y_i=1}X_i\|_{\infty}$. The intercept term
is dropped in this setting for simplicity, which implies that
$P(Y=1)=P(Y=-1)=0.5$, as assumed in our simulation setting. The
iteration scheme is detailed in Algorithm ~\ref{AlgoSparse} in
supplementary material. The loss decay plot is in Figure~\ref{LogisS},
which shows that our algorithm still outperforms other
algorithms. Note that SVRG-ADMM also performs without setting
$\alpha$, $\gamma$, $S$, $T$ in this case, however, here these
parameters are easy to set and SS-PRSM seems to be robust to their specification.

\subsection{Real Data}

We investigate four real datasets to illustrate the efficiency of our
algorithm: i) communities and crime data set~\footnote{Download from
  http://archive.ics.uci.edu/ml/datasets/communities+and+crime.},
EE2006-tfidf~\footnote{Download from
  https://www.csie.ntu.edu.tw/~cjlin/libsvmtools/datasets/regression.html.},
sido~\footnote{Download from
  http://www.causality.inf.ethz.ch/data/SIDO.html.}, and NSL-KDD data
set~\footnote{Details in http://www.unb.ca/cic/datasets/index.html and
  download from https://github.com/defcom17/NSL\_KDD.}.
Table~\ref{RealData} summarizes different setting and parameters used.
We briefly introduce the last NSL-KDD data. Recently, machine learning
methods have been widely used in abnormal flow detection, which is
increased dramatically with upgrading the computer network, hardware
and software.  ADMM based framework has a good parallel performance to
effectively cope with large data (see~\cite{bamakan2017ramp}). We are
trying to test the efficiency of our proposed algorithm (which is PRSM
based). We fit a sparse logistic for KDD because of $p$ is large and
our response is a binary variable characterizing the connection type:
normal(1), doubtful(0).

\begin{figure}[h]
	\centering
	\begin{subfigure}[b]{0.4\textwidth}
		\includegraphics[width=\textwidth]{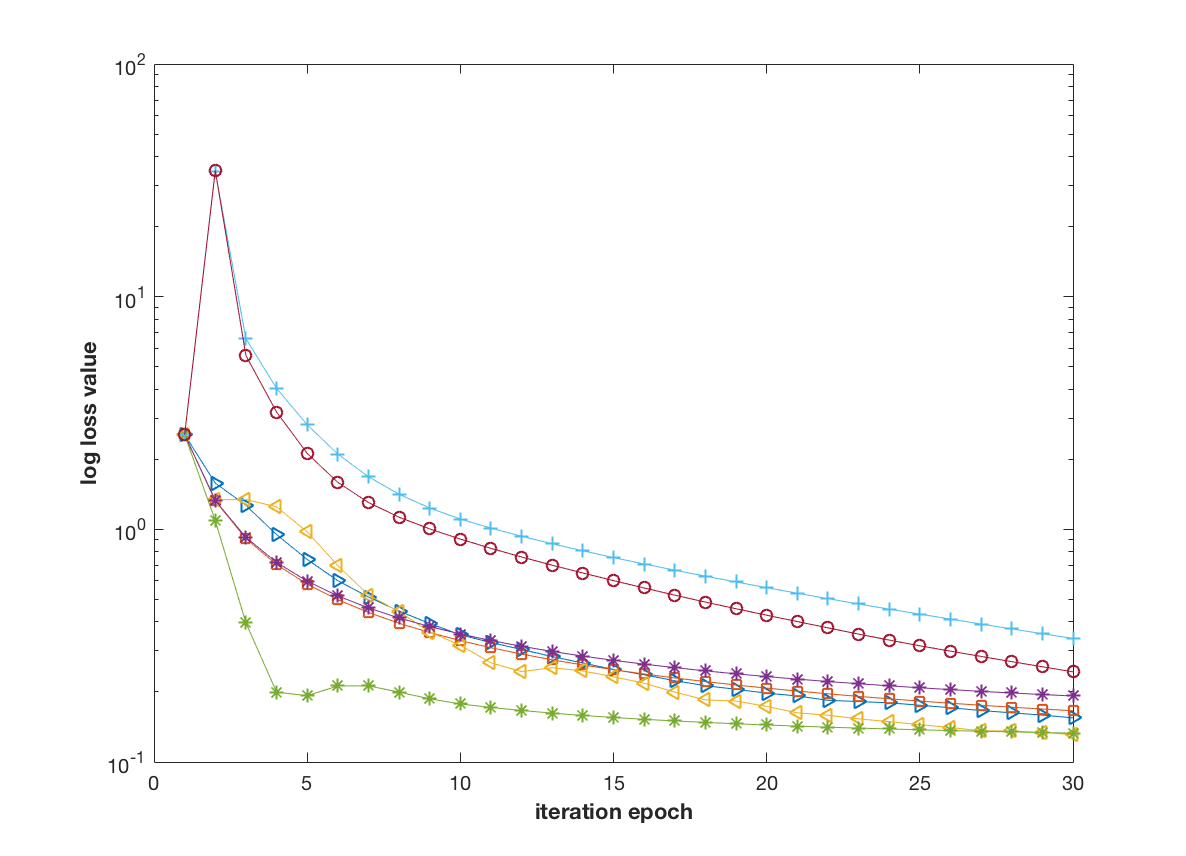}
		\caption{crime}
		\centering
		\label{Lasso3}
	\end{subfigure}
	~
	\begin{subfigure}[b]{0.4\textwidth}
		\includegraphics[width=\textwidth]{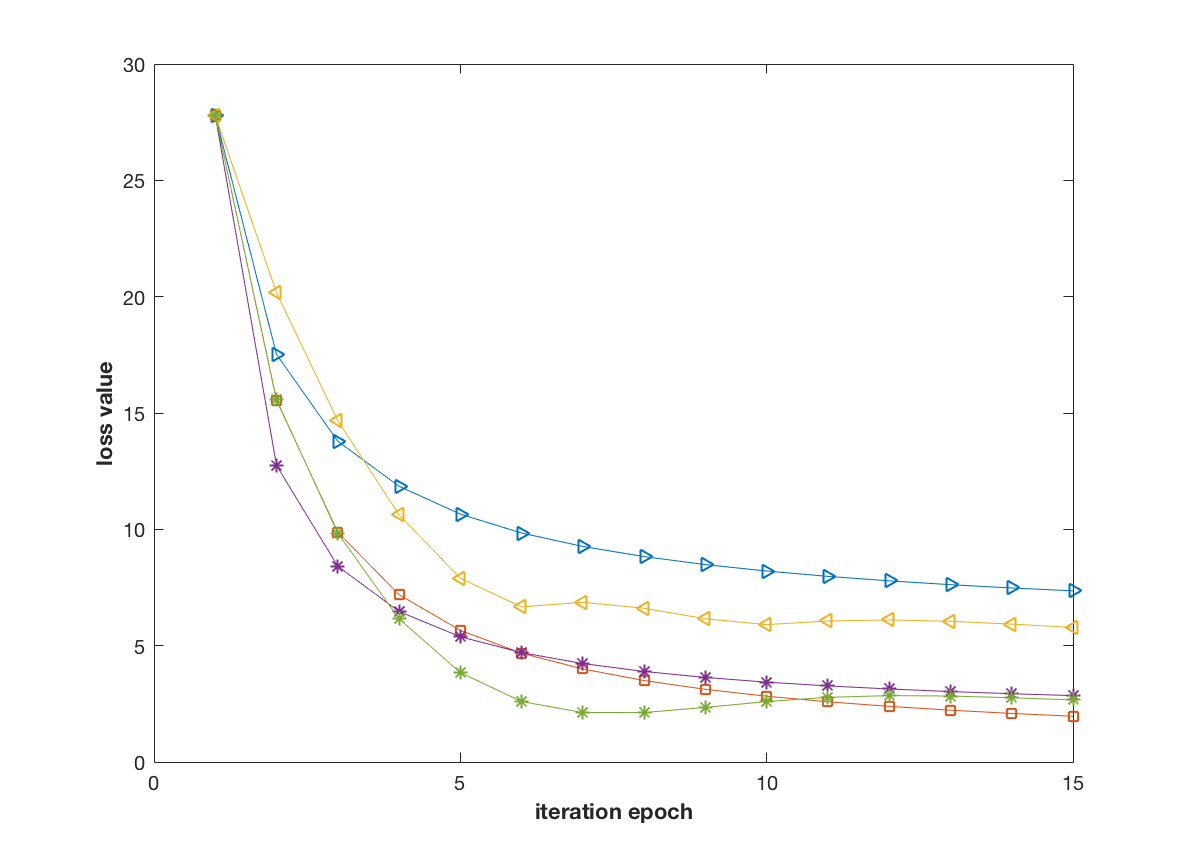}
		\caption{E2006}\centering
		\label{Lasso4}
		\centering
	\end{subfigure}
	~
	\begin{subfigure}[b]{0.4\textwidth}
		\includegraphics[width=\textwidth]{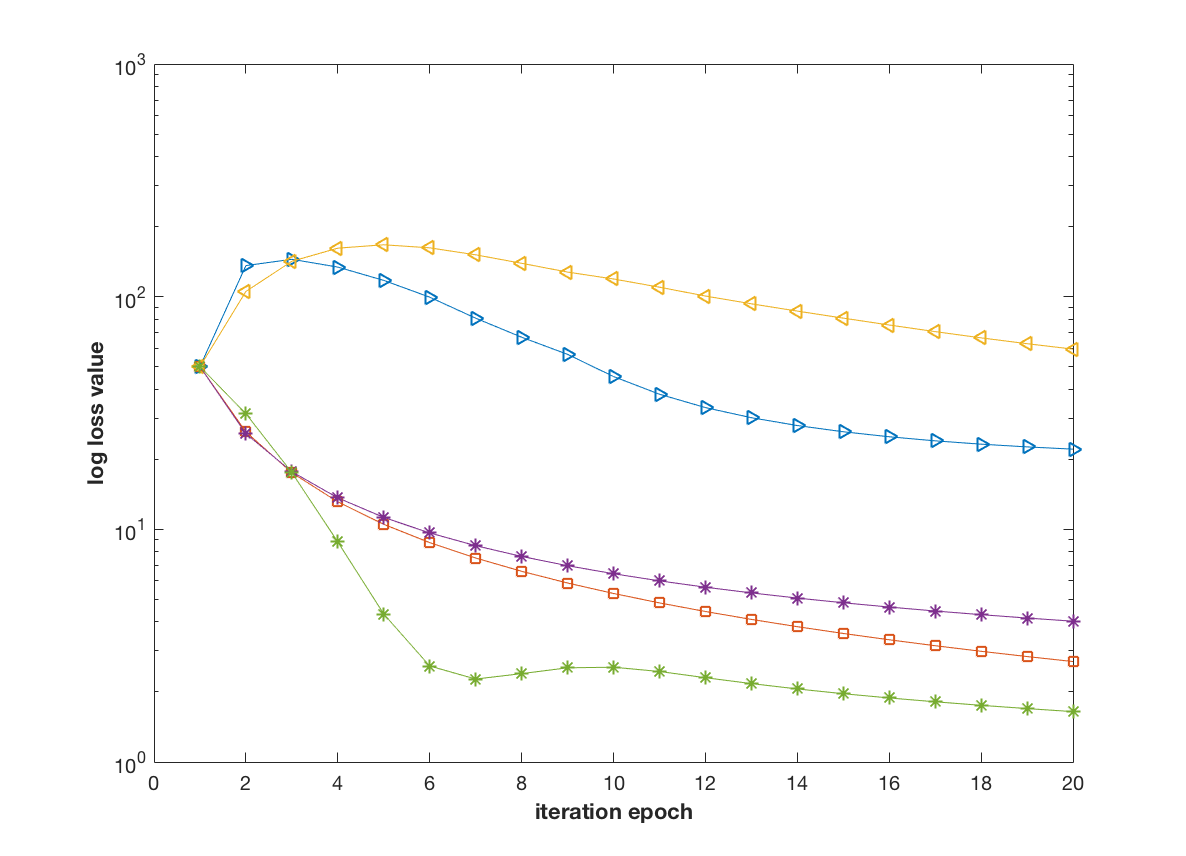}
		\caption{sido}\centering
		\label{Logis1}
		\centering
	\end{subfigure}
	~
	\begin{subfigure}[b]{0.4\textwidth}
		\includegraphics[width=\textwidth]{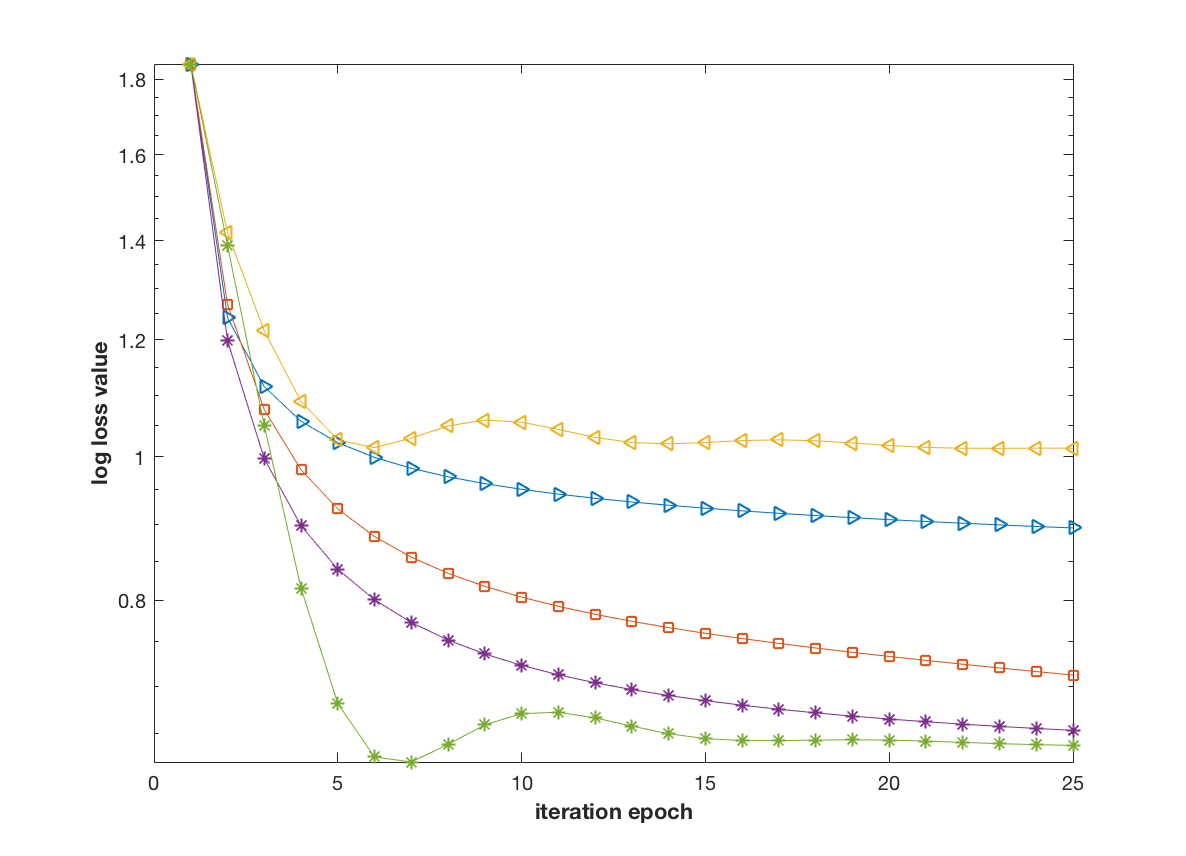}
		\caption{KSL-KDD}\centering
		\label{Logis2}
		\centering
	\end{subfigure}
	\includegraphics[width=15cm,height=0.27cm]{legend.jpeg}
	\caption{\textit{This figure shows loss decay trajectories on
            real data sets. The left two plots correspond to fitting a
            lasso model, while the right two plots correspond to
            sparse logistic regression. Because SC-PRSM and ADMM need
            to compute the inverse of a huge matrix $(\beta I+X^TX)$,
            we did not implement them in the case of E2006.}}\label{fig:2}
\end{figure}

The loss decay plots are in Figure ~\ref{fig:2}. SSPB-SCPRSM and
S-ADMM always converge slower than other stochastic
algorithms. SVRG-ADMM and SCAS-ADMM are similar, while our SS-PRSM can
converge faster than them.
\begin{table}[h!]
	\centering
	\begin{tabular}{ |c|c|c|c|c|c|c|c|c|c|c|c| }
		\hline
		Data  & Model & $n$ & $p$ & $\zeta$ & $\epsilon$ &   $\beta$&$\alpha$& $\gamma$ &$S$&$T$\\
		\hline
		crime & Lasso & 1994 & 122 & 0.02 & $10^{-11}$ & 5 &0.8 & 0.3& $2I$ &$2.5I$ \\
		\hline
		E2006-tfidf & Lasso & 16,087 &  150,360 & 0.0001 &$10^{-11}$ & 0.8& 0.9&0.1 &$I$ & $1.8I$\\
		\hline
		sido & Sparse Logistic & 12,678 &4,932 & 0.01 & $10^{-8}$ & 1 &0.5 &0.3 & $I$ & $I$\\
		\hline
		KSL-KDD & Sparse Logistic & 125973 & 115 & 0.01 & $10^{-11}$ &1 &0.9&0.3&$I$ & $2I$\\
		\hline
	\end{tabular}
\caption{Real Data Statistic}\label{RealData}
\end{table}

\section{Conclusion}\label{Con}

In this paper, we propose a new stochastic algorithm, SS-PRSM. The
resulted algorithm has $O(1/K)$ ergodic convergence rate, which
matches the rate of many start-of-the-art variants of stochastic
ADMM. We bridge the ergodic convergence rate gap between batch and
stochastic algorithms for PRSM theoretically. Furthermore, we show
that our algorithm outperforms ADMM based algorithms when fitting 
statistical models such as Lasso, group Lasso, and sparse logistic
regression.

The theoretical analysis of the convergence rate in non-ergodic sense
for SS-PRSM is still an open problem and is much more difficult than
for an ADMM based algorithm because the adjustment of the iteration
scheme. Meanwhile, we can also try to incorporate other tricks,
such as conjugate gradient and Nesterov's extrapolation, for the
iteration of $\bx_1$. We think they can also attain $O(1/K)$ ergodic
convergence rate. On the other hand, the effectiveness of PRSM under
nonconvex setting, for both batch and stochastic version, is still
blank. SVRG-ADMM has good convergence properties under nonconvex
setting as shown in ~\cite{zheng2016fast}. However, the general
theoretical convergence result for PRSM is still unknown, which is
also an interesting area.

\newpage
\bibliography{reference}
\bibliographystyle{plain}

\newpage
\section*{\centering{Supplementary Materials:\\Stochastic Approaches for Peaceman-Rachford Splitting Method }}

\subsection*{$\bullet$ Proof of Lemma 4.1}
We only consider one more step after getting $\bx_1^k, \bx_2^k, \blambda^k$. Based on our Algorithm 3, $\forall \bx_1\in\X_1$, we have 
\begin{align*}\label{equ1}
\|\bx_{1,t+1}-\bx_1\|_2^2&=\|\bx_{1,t}-\eta_k\nabla g_{\beta}(\bx_{1,t},\xi_{k+1}^{t+1})-\bx_1\|^2\\
&=\|\bx_{1,t}-\bx_1\|^2-2\eta_k\langle \nabla g_{\beta}(\bx_{1,t},\xi_{k+1}^{t+1}), \bx_{1,t}-\bx_1\rangle+\eta_k^2\|\nabla g_{\beta}(\bx_{1,t},\xi_{k+1}^{t+1})\|^2. \tag 1
\end{align*}
Let's deal with $\langle \nabla g_{\beta}(\bx_{1,t},\xi_{k+1}^{t+1}), \bx_{1,t}-\bx_1\rangle$ first. Note that
\begin{align*}\label{equ2}
\nabla g_{\beta}(\bx_{1,t},\xi_{k+1}^{t+1})&=(G_{\beta}^{k})^{\prime}(\bx_{1,t},\xi^{t+1}_{k+1})-(G_{\beta}^{k})^{\prime}(\tilde{\bx}_1,\xi^{t+1}_{k+1})+\tilde{\mu}\\
&=\theta_{1,\xi^{t+1}_{k+1}}'(\bx_{1,t})-A^T\blambda^k+\beta A^T(A\bx_{1,t}+B\bx_2^k-\bb)+S(\bx_{1,t}-\bx_1^k)-\theta_{1,\xi^{t+1}_{k+1}}'(\tilde{\bx}_1)+\theta_{1}'(\tilde{\bx}_1)\\
&=\theta_1'(\bx_{1,t})-A^T\blambda^k+\beta A^T(A\bx_{1,t}+B\bx_2^k-\bb)+S(\bx_{1,t}-\bx_1^k)+\delta_{\xi_{k+1}^{t+1}}(\bx_{1,t})-\delta_{\xi_{k+1}^{t+1}}(\tilde{\bx}_{1}), \tag 2
\end{align*}
where $\delta_i(\bx_1)=\theta'_{1i}(\bx_1)-\theta'_{1}(\bx_1)$ $\forall \bx_1\in\X_1$. So we know
\begin{align*}\label{equ3}
\mE \delta_i(\bx_1)=0\ \ \ \ \forall \bx_1\in\X_1, \forall i=1,2,...,n. \tag 3
\end{align*}
Go back to equation (~\ref{equ1}) and plug in equation (~\ref{equ2}), we have
\begin{align*}
\|\bx_{1,t}-\bx_1\|^2-\|\bx_{1,t+1}&-\bx_1\|^2+\eta_k^2\|\nabla g_{\beta}(\bx_{1,t},\xi_{k+1}^{t+1})\|^2=2\eta_k\langle \nabla g_{\beta}(\bx_{1,t},\xi_{k+1}^{t+1}), \bx_{1,t}-\bx_1\rangle\\
=&2\eta_k\langle\theta_1'(\bx_{1,t})-A^T\blambda^k+\beta A^T(A\bx_{1,t}+B\bx_2^k-\bb)+S(\bx_{1,t}-\bx_1^k), \bx_{1,t}-\bx_1\rangle\\
&+\langle \delta_{\xi_{k+1}^{t+1}}(\bx_{1,t})-\delta_{\xi_{k+1}^{t+1}}(\tilde{\bx}_{1}), \bx_{1,t}-\bx_1\rangle\\
\geq&2\eta_k(\theta_1(\bx_{1,t})-\theta_1(\bx_1))+\langle S(\bx_{1,t}-\bx_1^k)-A^T\blambda^k+\beta A^T(A\bx_{1,t}+B\bx_2^k-\bb), \bx_{1,t}-\bx_1\rangle\\
&+\underbrace{\langle \delta_{\xi_{k+1}^{t+1}}(\bx_{1,t})-\delta_{\xi_{k+1}^{t+1}}(\tilde{\bx}_{1}), \bx_{1,t}-\bx_1\rangle}_{\Delta(k,t)}. 
\end{align*}
Sum over $t=0,1,...,M_k-1$ both side, we can get
\begin{align*}\label{equ4}
\|\bx_{1,0}-\bx_1\|^2-&\|\bx_{1,M_k}-\bx_1\|^2+\eta_k^2\sum_{t=0}^{M_k-1}\|\nabla g_{\beta}(\bx_{1,t},\xi_{k+1}^{t+1})\|^2\geq 2\eta_k M_k(\frac{1}{M_k}\sum_{t=0}^{M_k-1}\theta_1(\bx_{1,t})-\theta_1(\bx_1))\\
+&2\eta_k\bigg(\sum_{t=0}^{M_k-1}\langle S(\bx_{1,t}-\bx_1^k)-A^T\blambda^k+\beta A^T(A\bx_{1,t}+B\bx_2^k-\bb), \bx_{1,t}-\bx_1\rangle+\sum_{t=0}^{M_k-1}\Delta(k,t)\bigg)\\
\geq &2\eta_kM_k(\theta_{1}(\bx_1^{k+1})-\theta_1(\bx_1))+2\eta_k\sum_{t=0}^{M_k-1}\Delta(k,t)\\
&+2\eta_kM_k\langle S(\bx_{1}^{k+1}-\bx_1^k)-A^T\blambda^k+\beta A^T(A\bx_{1}^{k+1}+B\bx_2^k-\bb), \bx_{1}^{k+1}-\bx_1\rangle. \tag 4
\end{align*}
The second inequality is due to (i) $\bx_1^{k+1}=\frac{1}{M_k}\sum_{t=0}^{M_k-1}\bx_{1,t}$; (ii) $\theta_1(\cdot)$ is convex function; (iii) $\langle (S+\beta A^TA)\bx_{1,t}, \bx_{1,t}-\bx_1\rangle$ is convex ($S\succeq0$). So, further from equation (~\ref{equ4}) we have
\begin{align*}\label{equ5}
\theta_1(\bx_1)-\theta_1(\bx_1^{k+1})&-\frac{1}{M_k}\sum_{t=0}^{M_k-1}\Delta(k,t)+\frac{D_{\X_1}^2}{2\eta_kM_k}+\frac{\eta_k}{2M_k}\sum_{t=0}^{M_k-1}\|\nabla g_{\beta}(\bx_{1,t},\xi_{k+1}^{t+1})\|^2\\
&+\langle \bx_1-\bx_1^{k+1}, S(\bx_{1}^{k+1}-\bx_1^k)-A^T\blambda^k+\beta A^T(A\bx_{1}^{k+1}+B\bx_2^k-\bb)\rangle\geq 0. \tag 5
\end{align*}
\qed 

\subsection*{$\bullet$ Proof of Lemma 4.2}
For $\forall k=0,1,2,...$ and $\forall t=0,1,...,M_k-1$, we can get
\begin{align*}\label{equ6}
\mE [\|\nabla g_{\beta}(\bx_{1,t},\xi_{k+1}^{t+1})\|^2]=&\mE[\|(G_{\beta}^{k})^{\prime}(\bx_{1,t},\xi^{t+1}_{k+1})-(G_{\beta}^{k})^{\prime}(\tilde{\bx}_1,\xi^{t+1}_{k+1})+\tilde{\mu}\|^2]\\
\leq & 2\mE[\|G_{\beta}^{k})^{\prime}(\bx_{1,t},\xi^{t+1}_{k+1})-(G_{\beta}^{k})^{\prime}(\tilde{\bx}_1,\xi^{t+1}_{k+1})\|^2]+2\|\tilde{\mu}\|^2\\
= & 2\mE[\|\theta_{1,\xi_{k+1}^{t+1}}'(\bx_{1,t})-\theta_{1,\xi_{k+1}^{t+1}}'(\tilde{\bx}_{1})+(\beta A^TA+S)(\bx_{1,t}-\tilde{\bx}_1)\|^2]+2\|\tilde{\mu}\|^2\\
=&\frac{2}{n}\sum_{i=1}^{n}\|\theta_{1i}'(\bx_{1,t})-\theta_{1i}'(\tilde{\bx}_{1})+(\beta A^TA+S)(\bx_{1,t}-\tilde{\bx}_1)\|^2+2\|\tilde{\mu}\|^2. \tag 6
\end{align*}
Note that
\begin{align*}\label{equ7}
\|\theta_{1i}'(\bx_{1,t})-\theta_{1i}'(\tilde{\bx}_{1})+(\beta A^TA+S)(\bx_{1,t}-\tilde{\bx}_1)\|&\leq(\nu_i+\beta\sigma_A^2+\sigma_S)\|\bx_{1,t}-\tilde{\bx}_1\|\\
&\leq(\nu_i+\beta\sigma_A^2+\sigma_S)D_{\X_1}. \tag 7
\end{align*}
where $\sigma_A$ and $\sigma_S$ are singular value of $A$ and $S$ respectively. Plug (~\ref{equ7}) into (~\ref{equ6}) and we have
\begin{align*}\label{equ8}
\mE [\|\nabla g_{\beta}(\bx_{1,t},\xi_{k+1}^{t+1})\|^2]&\leq\frac{2D^2_{\X_1}}{n}\sum_{i=1}^{n}(\nu_i+\beta\sigma_A^2+\sigma_S)^2+2\|\tilde{\mu}\|^2\\
&\leq 2D^2_{\X_1}(\nu+\beta\sigma^2_A+\sigma_S)^2+2\|\tilde{\mu}\|^2. \tag 8
\end{align*}
Because $\|\tilde{\mu}\|=\|(G_{\beta}^{k})^{\prime}(\bx_1^k)\|=C_k$, we get final argument.

\qed

\subsection*{$\bullet$ Proof of Theorem 1}
Because our iteration scheme for $\bx_2$ and $\blambda$ are the same as  Na et al. (2016), we can directly use their results with small modification to get following argument. But we still show how to prove the argument after this main theorem for self-contain.  

\noindent\textit{\textbf{Argument:}} \label{Arg}

Let the sequence $\{\bw^k\}_{k=1}^{\infty}$ be generated by SS-PRSM. If $\alpha\in[0,1)$ and $\gamma\in(0,\frac{1-\alpha+\sqrt{(1+\alpha)^2+4(1-\alpha^2)}}{2})$, then $\forall \bw\in \Omega$ we have
\begin{align*}
\theta(\bu^{k+1})-\theta(\bu)+\langle \bw^{k+1}-\bw, F(\bw^{k+1})\rangle\leq& -\frac{1}{M_k}\sum_{t=0}^{M_k-1}\Delta(k,t)+\frac{D_{\X_1}^2}{2\eta_kM_k}+\frac{\eta_k}{2M_k}\sum_{t=0}^{M_k-1}\|\nabla g_{\beta}(\bx_{1,t},\xi_{k+1}^{t+1})\|^2\\
&+\frac{1}{2}(\|\bw^k-\bw\|^2_G+\zeta_{\alpha}\|\bx_2^k-\bx_2^{k-1}\|_T^2+\rho_{\alpha,\gamma}\beta\|\br^k\|^2)\\
&-\frac{1}{2}(\|\bw^{k+1}-\bw\|^2_G+\zeta_{\alpha}\|\bx_2^{k+1}-\bx_2^{k}\|_T^2+\rho_{\alpha,\gamma}\beta\|\br^{k+1}\|^2),
\end{align*}
where $\zeta_\alpha, \rho_{\alpha,\gamma}\geq 0$, $\br^{k}=A\bx_1^{k}+B\bx_2^k-\bb$ for $k=1,2,...$, and $G
=\left( \begin{array}{ccc}
S & 0 &0 \\
0 & T+\frac{\alpha+\gamma-\alpha\gamma}{\alpha+\gamma}\beta B^{T}B & -\frac{\alpha}{\alpha+\gamma}B^T \\
0 & -\frac{\alpha}{\alpha+\gamma}B & \frac{1}{(\alpha+\gamma)\beta}I_m \end{array} \right)$.

\noindent It's easy to show that $\langle \bw_1-\bw_2, F(\bw_1)\rangle=\langle \bw_1-\bw_2, F(\bw_2)\rangle$ for $\forall \bw\in\Omega$. Combine this with above argument and we have $\forall \bw\in\Omega$
\begin{align*}\label{equ9}
\theta(\bar{\bu}_K)-\theta(\bu)+\langle \bar{\bw}_K-\bw, F(\bw)\rangle\leq&
\frac{1}{K}\sum_{k=1}^{K}\bigg(\theta(\bu^{k+1})-\theta(\bu)+\langle \bw^{k+1}-\bw, F(\bw)\rangle\bigg)\\
\leq& -\frac{1}{K}\sum_{k=1}^{K}\frac{1}{M_k}\sum_{t=0}^{M_k-1}\Delta(k,t)+\frac{1}{K}\sum_{k=1}^{K}\bigg(\frac{D_{\X_1}^2}{2\eta_kM_k}+\frac{\eta_k}{2M_k}\sum_{t=0}^{M_k-1}\|\nabla g_{\beta}(\bx_{1,t},\xi_{k+1}^{t+1})\|^2\bigg)\\
&+\frac{1}{2K}(\|\bw^1-\bw\|^2_G+\zeta_{\alpha}\|\bx_2^1-\bx_2^{0}\|_T^2+\rho_{\alpha,\gamma}\beta\|\br^1\|^2). \tag 9
\end{align*}
So, taking expectation on both side and combining with Lemma 4.2, we can get
\begin{align*}
\mE[\theta(\bar{\bu}_K)-\theta(\bu)+\langle \bar{\bw}_K-\bw, F(\bw)\rangle]\leq& \frac{1}{2K}(\|\bw^1-\bw\|^2_G+\zeta_{\alpha}\|\bx_2^1-\bx_2^{0}\|_T^2+\rho_{\alpha,\gamma}\beta\|\br^1\|^2)\\
&+\frac{1}{K}\sum_{k=1}^{K}\bigg(\frac{D_{\X_1}^2}{2\eta_kM_k}+\eta_k(\underbrace{D^2_{\X_1}(\nu+\beta\sigma^2_A+\sigma_S)^2}_{C^2}+C_k^2)\bigg).
\end{align*}
Here, we can choose $\eta_k$ and $M_k$ to make right-hand side has $O(1/T)$. For example, let
\begin{align*}\label{equ10}
\eta_k&=\frac{1}{(C^2+C_k^2)(k+1)^2},\\
M_k&=C^2+C_k^2, \tag {10}
\end{align*}
and we get $\mE[\theta(\bar{\bu}_K)-\theta(\bu)+\langle \bar{\bw}_K-\bw, F(\bw)\rangle]\approx O(\frac{1}{K})$ for $\forall \bw\in\Omega$.

\noindent Last we will get another convergence rate for another criteria. We go back to the left-hand side of equation ~\ref{equ9}. Because of randomness of $\bw$, we plug in $\tilde{\bw}=(\bx_1^*, \bx_2^*, \blambda)$ where $\blambda\in\mathcal{B}_{\rho}=\{\blambda, \|\blambda\|_2\leq\rho\}$ for any $\rho>0$. We get
\begin{align*}
\theta(\bar{\bu}_K)-\theta(\bu^*)+\langle \bar{\bw}_K-\tilde{\bw}, F(\tilde{\bw})\rangle=&\theta(\bar{\bu}_K)-\theta(\bu^*)-\blambda(A\bar{\bx}_{1K}+B\bar{\bx}_{2K}-\bb).
\end{align*}
So, we have $\forall\blambda\in \mathcal{B}_{\rho}$,
\begin{align*}
\theta(\bar{\bu}_K)-\theta(\bu^*)-\blambda(A\bar{\bx}_{1K}+B\bar{\bx}_{2K}-\bb)\leq
&-\frac{1}{K}\sum_{k=1}^{K}\frac{1}{M_k}\sum_{t=0}^{M_k-1}\Delta(k,t)+\frac{1}{K}\sum_{k=1}^{K}\bigg(\frac{D_{\X_1}^2}{2\eta_kM_k}+\frac{\eta_k}{2M_k}\sum_{t=0}^{M_k-1}\|\nabla g_{\beta}(\bx_{1,t},\xi_{k+1}^{t+1})\|^2\bigg)\\
&+\frac{1}{2K}(\|\bw^1-\bw\|^2_G+\zeta_{\alpha}\|\bx_2^1-\bx_2^{0}\|_T^2+\rho_{\alpha,\gamma}\beta\|\br^1\|^2).
\end{align*}
Take supremum over $\mathcal{B}_{\rho}$ on both side,
\begin{align*}
\theta(\bar{\bu}_K)-\theta(\bu^*)&+\rho\|A\bar{\bx}_{1K}+B\bar{\bx}_{2K}-\bb\|_2\leq
-\frac{1}{K}\sum_{k=1}^{K}\frac{1}{M_k}\sum_{t=0}^{M_k-1}\Delta(k,t)+\frac{1}{K}\sum_{k=1}^{K}\bigg(\frac{D_{\X_1}^2}{2\eta_kM_k}+\frac{\eta_k}{2M_k}\sum_{t=0}^{M_k-1}\|\nabla g_{\beta}(\bx_{1,t},\xi_{k+1}^{t+1})\|^2\bigg)\\
&+\frac{1}{2K}\bigg(\|\bu^1-\bu\|^2_L+\zeta_{\alpha}\|\bx_2^1-\bx_2^{0}\|_T^2+\rho_{\alpha,\gamma}\beta\|\br^1\|^2+\frac{2\alpha\sigma_B\|\bx_2^1-\bx_2\|(\|\blambda^1\|+\rho)}{\alpha+\gamma}+\frac{2(\|\blambda^1\|^2+\rho^2)}{(\alpha+\gamma)\beta}\bigg),
\end{align*}
where $\sigma_B=\|B\|_2$ and $L=\begin{pmatrix}
S & 0 \\
0 & T+\frac{\alpha+\gamma-\alpha\gamma}{\alpha+\gamma}\beta B^{T}B
\end{pmatrix}$.\\
By taking expectation on both side and setting $\eta_k$ and $M_k$ as eqation ~\ref{equ10}, we can get $\forall \rho>0$
\begin{align*}
\mE[\theta(\bar{\bu}_K)-\theta(\bu^*)+\rho\|A\bar{\bx}_{1K}+B\bar{\bx}_{2K}-\bb\|_2]\approx O(\frac{1}{K}).
\end{align*}

\qed

\subsection*{$\bullet$ Iteration scheme for Lasso and sparse logistic regression}
Our iteration scheme for Lasso is given by following Algorithm ~\ref{SVRGLasso}. For sparse logistic regression, we adjust 3rd-14th rows in Algorithm ~\ref{SVRGLasso} to following Algorithm ~\ref{AlgoSparse}.
\begin{algorithm}[H]
	\caption{SS-PRSM for Lasso}
	\label{SVRGLasso}
	\begin{algorithmic}[1]
\State Input: $Y$, $X$, $\zeta$, $n$, $p$, ;
\State Specify $\epsilon$ (threshold), $\beta$, $\alpha$, $\gamma$, $S$, $T=aI$ (for some $a$);
\State $\nu=2\max_i\|X_iX_i^T\|_2$, $\sigma_S=\|S\|_2$;
\State $t=0$, $Z_1^0=0$, $Z_2^0=1$, $\blambda^0=0$, $C=p(\nu+\beta+\sigma_S)$;
\While {$\|Z_1^t-Z_2^t\|_2>\epsilon$}
\State $\tilde{Z}_1=Z_1^t$, $Z_{1,0}=\tilde{Z}_1$; \Comment{iteration for the first variable}
\State $\tilde{\mu}=(\beta I+\frac{2}{n}\sum_{i=1}^{n}X_iX_i^T)\tilde{Z}_1-\frac{2}{n}\sum_{i=1}^{n}Y_iX_i-\beta Z_2^t-\blambda^t$;
\State $C_t=\|\tilde{\mu}\|_2$, $M_t=\lceil C^2+C_t^2\rceil$, $\eta_t=\frac{1}{M_t(t+1)^2}$;
\For {$i=1,2,...,M_t-1$}
\State $\xi\sim\U\{1,n\}$;
\State $d_i=(S+\beta I+2X_{\xi}X_{\xi}^T)(Z_{1,i-1}-\tilde{Z}_{1})+\tilde{\mu}$;
\State $Z_{1,i}=Z_{1,i-1}-\eta_td_i$
\EndFor
\State $Z_1^{t+1}=\frac{1}{M_t}\sum_{i=0}^{M_t-1}Z_{1,i}$;
\State $\blambda^{t+1/2}=\blambda^t-\alpha\beta(Z_1^{t+1}-Z_2^t)$;
\State $Z_2^{t+1}=Soft_{\frac{\zeta}{a+\beta}}(\frac{aZ_2^t+\beta Z_1^{t+1}-\blambda^{t+1/2}}{a+\beta})$; \Comment{iteration for the second variable}
\State $\blambda^{t+1}=\blambda^{t+1/2}-\gamma\beta(Z_1^{t+1}-Z_2^{t+1})$;
\State $t=t+1$;
\EndWhile
\State Output: $Z_1=\frac{1}{t+1}\sum_{i=0}^{t}Z_1^i$
	\end{algorithmic}
\end{algorithm}

\begin{algorithm}[H]
	\caption{SS-PRSM for sparse logistic regression}
	\label{AlgoSparse}
	\begin{algorithmic}[1]
\State $\nu=\max_iY_i^2\|X_i\|^2_2$, $\sigma_S=\|S\|_2$;
\State $t=0$, $Z_1^0=0$, $Z_2^0=1$, $\blambda^0=0$, $C=p(\nu+\beta+\sigma_S)$;
\While {$\|Z_1^t-Z_2^t\|_2>\epsilon$}
\State $\tilde{Z}_1=Z_1^t$, $Z_{1,0}=\tilde{Z}_1$; \Comment{iteration for the first variable}
\State $\tilde{\mu}=\beta \tilde{Z}_1-\frac{1}{n}\sum_{i=1}^{n}\frac{Y_iX_i}{1+\exp(Y_iX_i^T\tilde{Z}_1)}-\beta Z_2^t-\blambda^t$;
\State $C_t=\|\tilde{\mu}\|_2$, $M_t=C^2+C_t^2$, $\eta_t=\frac{1}{M_t(t+1)^2}$;
\For {$i=1,2,...,M_t-1$}
\State $\xi\sim\U\{1,n\}$;
\State $v_1=Y_{\xi}X_{\xi}^TZ_{1,i-1}$, $v_2=Y_{\xi}X_{\xi}^T\tilde{Z}_{1}$;
\State $d_i=(\beta I+S)(Z_{1,i-1}-\tilde{Z}_1)+\frac{(e^{v_1}-e^{v_2})Y_{\xi}X_{\xi}}{(1+e^{v_1})(1+e^{v_2})}+\tilde{\mu}$;
\State $Z_{1,i}=Z_{1,i-1}-\eta_td_i$
\EndFor
\State $Z_1^{t+1}=\frac{1}{M_t}\sum_{i=0}^{M_t-1}Z_{1,i}$;
\State 15th-17th rows in Algorithms ~\ref{SVRGLasso};
\EndWhile
\State Output: $Z_1$
	\end{algorithmic}
\end{algorithm}

\newpage
\subsection*{$\bullet$ Proof of Argument}
We will mainly prove how to get the Argument in the Proof of Theorem 1. It just follows the work in Gu et al. (2015) and Na et al. (2016) with some small modifications. But for completeness, we still prove the argument.

\noindent$\bullet$\textit{\textbf{Notations:}}\\
We first propose some notations that will only be used in this appendix. They are coincident with Na et al (2016).
\begin{itemize}
	\item $H=\frac{1}{\alpha+\gamma}\left( \begin{array}{ccc}
		(\alpha+\gamma-\alpha\gamma)\beta B^{T}B & -\alpha B^T \\
		-\alpha B & \frac{1}{\beta}I_m \end{array} \right)$; $\bv=\begin{pmatrix}
		\bx_2\\
		\blambda
		\end{pmatrix}$
\end{itemize}
\noindent$\bullet$\textit{\textbf{Proof:}}

\noindent Based on the iteration scheme for $\bx_2$, we have $\forall \bx_2\in\X_2$,
\begin{align*}\label{equ1}
\theta_2(\bx_2)-\theta_2(\bx^{k+1}_2)+\langle \bx_2-\bx_2^{k+1},T(\bx_2^{k+1}-\bx_2^{k})-B^T\blambda^{k+1/2}+\beta B^T(A\bx_1^{k+1}+B\bx_2^{k+1}-\bb) \geq 0. \tag 1
\end{align*}
Further, we simplify the iteration for $\blambda^{k+1/2}, \blambda^{k+1}$, and have
\begin{align*}
\blambda^{k}&=\blambda^{k+1/2}+\alpha\beta(A\bx_1^{k+1}+B\bx_2^k-\bb)\\
&=\blambda^{k+1}+\gamma\beta\br^{k+1}+\alpha\beta\br^{k+1}+\alpha\beta B(\bx_2^k-\bx_2^{k+1})\\
&=\blambda^{k+1}+(\alpha+\gamma)\beta\br^{k+1}+\alpha\beta B(\bx_2^k-\bx_2^{k+1}).\\
\end{align*} 
So, we get
\begin{align*}\label{equ2}
\br^{k+1}-\frac{\alpha}{\alpha+\gamma}B(\bx_2^{k+1}-\bx_2^k)+\frac{1}{(\alpha+\gamma)\beta}(\blambda^{k+1}-\blambda^k)=0. \tag 2
\end{align*}
If we define $P_k=\theta(\bu)-\theta(\bu^{k+1})-\frac{1}{M_k}\sum_{t=0}^{M_k-1}\Delta(k,t)+\frac{D_{\X_1}^2}{2\eta_kM_k}+\frac{\eta_k}{2M_k}\sum_{t=0}^{M_k-1}\|\nabla g_{\beta}(\bx_{1,t},\xi_{k+1}^{t+1})\|^2$, 
then based on the Lemma 4.1 and above equation ~\ref{equ1}, \ref{equ2}, we will get $\forall \bw\in \Omega$,
\begin{align*}\label{equ3}
P_k&+\underbrace{\langle\bw-\bw^{k+1}, \left( \begin{array}{ccc}
	S(\bx_1^{k+1}-\bx_1^k) \\
	T(\bx_2^{k+1}-\bx_2^k) \\
	0  \end{array} \right)+\left( \begin{array}{ccc}
	0 \\
	\alpha\beta B^T\br^{k+1}+(1-\alpha)\beta B^T B(\bx_2^{k+1}-\bx_2^k) \\
	-\frac{\alpha}{\alpha+\gamma}B(\bx_2^{k+1}-\bx_2^k)+\frac{1}{(\alpha+\gamma)\beta}(\blambda^{k+1}-\blambda^k)  \end{array} \right)\rangle}_{Z_{k}}\\
&+\underbrace{\langle \bw-\bw^{k+1},\left( \begin{array}{ccc}
	A^T \\
	B^T \\
	0  \end{array} \right)\bigg((1-\alpha-\gamma)\beta\br^{k+1}+(1-\alpha)\beta B(\bx_2^k-\bx_2^{k+1})\bigg)\rangle}_{V_k} \geq \langle \bw^{k+1}-\bw, F(\bw^{k+1})\rangle. \tag 3 
\end{align*}
Based on simple algebra, we can show
\begin{itemize}
	\item $Z_k=\langle\bw-\bw^{k+1}, \left( \begin{array}{ccc}
	S(\bx_1^{k+1}-\bx_1^k) \\
	T(\bx_2^{k+1}-\bx_2^k) \\
	0  \end{array} \right)+\left( \begin{array}{ccc}
	0 \\
	H(\bv_{k+1}-\bv_k) \end{array} \right)\rangle=(\bw-\bw^{k+1})^TG(\bw^{k+1}-\bw^k)$;
	\item $
	V_k=\langle \br(\bw)-\br^{k+1}, (1-\alpha-\gamma)\beta\br^{k+1}+(1-\alpha)\beta B(\bx_2^k-\bx_2^{k+1})\rangle\\
	\text{\ \ \ \ }=-(1-\alpha-\gamma)\beta\|\br^{k+1}\|^2-(1-\alpha)\beta\langle \br^{k+1},B(\bx_2^k-\bx_2^{k+1})\rangle
	$.
\end{itemize}
Plug in equation ~\ref{equ3} and we get
\begin{align*}
P_k+(\bw-\bw^{k+1})^TG(\bw^{k+1}-\bw^k) \geq (1-\alpha-\gamma)\beta\|\br^{k+1}\|^2+(1-\alpha)\beta\langle \br^{k+1}, B(\bx_2^k-\bx_2^{k+1})\rangle+\langle \bw^{k+1}-\bw,F(\bw^{k+1})\rangle.
\end{align*}
Go through Lemma 2, Lemma 3, Lemma 4 in Na et al. (2016), we know 
when $\alpha\in [0, 1)$ and $\gamma \in (0,\frac{1-\alpha+\sqrt{(1+\alpha)^2+4(1-\alpha^2)}}{2})$, there exist $\zeta_\alpha, \rho_{\alpha,\gamma}\geq 0$ and $\tau_{\alpha,\gamma}\in(0,1)$ such that
$\forall \bw\in\Omega$, we have
\begin{align*}
2P_k&+(\|\bw^k-\bw\|_G^2+\zeta_\alpha\|\bx_2^k-\bx_2^{k-1}\|_T^2+\rho_{\alpha,\gamma}\beta\|\br^k\|^2)
-(\|\bw^{k+1}-\bw\|_G^2+\zeta_\alpha\|\bx_2^{k+1}-\bx_2^{k}\|_T^2+\rho_{\alpha,\gamma}\beta\|\br^{k+1}\|^2)\\
&\geq \tau_{\alpha,\gamma}\|\bw^k-\bw^{k+1}\|_G^2+2\langle\bw^{k+1}-\bw,F(\bw^{k+1})\rangle\geq2\langle\bw^{k+1}-\bw,F(\bw^{k+1})\rangle. 
\end{align*}

\qed

\end{document}